\documentclass{article}

\usepackage{arxiv}

\usepackage{newtxtext}
\usepackage{newtxmath}

\usepackage[utf8]{inputenc}
\usepackage[T2A,T1]{fontenc}
\usepackage[english]{babel}

\usepackage{hyperref}
\usepackage{url}
\usepackage{booktabs}
\usepackage{amsfonts}
\usepackage{nicefrac}
\usepackage{microtype}
\usepackage{cleveref}
\usepackage{graphicx}
\usepackage{natbib}
\usepackage{doi}
\usepackage{authblk}

\usepackage{orcidlink}

\title{A Comprehensive Analysis of Arabic Natural Language Processing Research: \\
       Trends, Topic Evolution, and Research Gaps -- \\
       A Bibliometric and Topic-Based Study}

\author{Mullosharaf ~K.~Arabov\thanks{Email: \texttt{MKArabov@kpfu.ru}}\orcidlink{0000-0003-2525-1183}}
\affil{Kazan Federal University, Institute of Computational Mathematics and Information Technologies, Kazan, Russia}

\renewcommand{\shorttitle}{Arabic NLP: A Bibliometric and Topic-Based Analysis}

\hypersetup{
    pdftitle={A Comprehensive Analysis of Arabic Natural Language Processing Research: Trends, Topic Evolution, and Research Gaps -- A Bibliometric and Topic-Based Study},
    pdfsubject={cs.CL, cs.AI, cs.DL},
    pdfauthor={M. K. Arabov},
    pdfkeywords={Arabic natural language processing, bibliometric analysis, topic modeling, BERTopic, research gaps, dialectal Arabic, large language models, sentiment analysis, named entity recognition, machine translation}
}

\begin{document}

\maketitle

\begin{abstract}
Arabic Natural Language Processing (NLP) has experienced rapid growth over the past decade, driven by digital transformation in the Arab world, the rise of social media, and recent breakthroughs in large language models (LLMs). Despite the proliferation of research, a comprehensive quantitative meta-analysis of the field remains absent. This study presents a large-scale bibliometric and topic-based analysis of 7,120 Arabic NLP papers published between 1960 and 2026, sourced from five platforms---arXiv, ACL Anthology, Semantic Scholar, Crossref, and OpenAlex---with an additional targeted OpenAlex subset. We employ BERTopic for transformer-based topic modeling, regression analysis to identify predictors of citation impact, social network analysis to map co-authorship structures, and geographic analysis to reveal regional research contributions.

Our findings show a significant acceleration in publication output, with the majority of papers published after 2020, driven by the rise of transformer-based models (AraBERT, MARBERT) and, more recently, LLMs. Topic modeling identifies 19 substantive research themes, with the largest cluster (2,942 papers) centered on text, speech, translation, and recognition. Citation analysis reveals a positive correlation between paper age and citations (r = 0.245, p < 0.001), while regression analysis (R² = 0.105) indicates that papers indexed in OpenAlex or Semantic Scholar and those with institutional affiliations receive significantly more citations on average (e.g., +11 citations for OpenAlex). Geographic analysis shows that Saudi Arabia, the United States, and Egypt lead in research output, with King Saud University (142 papers) and Cairo University (67 papers) as top contributors.

A task--dialect gap matrix identifies critical understudied areas, including summarization for Maghrebi (Darija), Iraqi, and Sudanese dialects, as well as dialect identification for Sudanese and Yemeni Arabic. The H-index analysis reveals that the largest topic (text, speech, translation, recognition) achieves the highest H-index of 90, followed by sentiment analysis with 57. We compare our findings with existing qualitative surveys, demonstrating that our quantitative approach complements them by providing empirical evidence for observed trends and gaps. This study offers actionable recommendations for researchers, funding agencies, and policymakers to prioritize under-resourced dialects, enhance reproducibility, and develop culturally aligned benchmarks for Arabic NLP.

\textbf{Keywords:} Arabic natural language processing, bibliometric analysis, topic modeling, BERTopic, research gaps, dialectal Arabic, large language models, sentiment analysis, named entity recognition, machine translation
\end{abstract}


\section{Introduction}\label{sec:intro}

Arabic is one of the most widely spoken languages in the world, with over 420 million native speakers across 22 countries and an additional 1.9 billion Muslims who use it as a liturgical language. Its linguistic complexity is considerable: a non-concatenative root-and-pattern morphology, a marked diglossia between Modern Standard Arabic (MSA) and numerous regional dialects, an orthography that often omits short vowels, and a right-to-left writing system. These features pose distinct challenges for computational modelling, and for many years they placed Arabic at the periphery of Natural Language Processing (NLP) research in comparison with English and other high-resource languages \citep{guellil2021arabic, alayba2025arabic, holes2004modern, watson2007phonology}.

The past decade, however, has witnessed a dramatic shift. The spread of social media has generated vast amounts of dialectal Arabic text, while national digital transformation agendas across the Arab world have created strong incentives for language technology. At the same time, the rise of pre-trained language models has begun to close the gap with English. Models such as AraBERT \citep{antoun2020arabert} and ARBERT/MARBERT \citep{abdulmageed2021arbert} have established new state-of-the-art results on a wide range of Arabic NLP tasks, including sentiment analysis, named entity recognition, machine translation, and question answering. Foundational tools such as Arabic morphological analyzers \citep{habash-rambow-2005-arabic,pasha-etal-2014-madamira} and pre-trained word embeddings \citep{soliman2017aravec} continue to underpin many of these applications. More recently, large language models (LLMs) such as Jais \citep{sengupta2023jais}, AceGPT \citep{huang2024acegpt}, and ALLaM \citep{saifulbari2025allam} have extended these capabilities to open-ended generation, instruction following, and reasoning. The field is therefore no longer limited by the absence of competitive models; the principal constraints are now the availability of high-quality datasets, the coverage of dialects, and the maturity of evaluation protocols.

This rapid expansion has been accompanied by a proliferation of survey articles that attempt to synthesise the state of the art. These surveys cover specific subfields: sentiment analysis \citep{shi2025comprehensive, amzil2025sentiment}, dialectal processing \citep{iwidat2026arabic, dahou2025survey, sakhi2026processing, aftan2025survey}, code-switching \citep{hamed2025survey}, evaluation of large language models \citep{alzubaidi2025evaluating, mashaabi2026survey, rhel2025large}, named entity recognition \citep{qu2024survey, elmoussaoui2024advancements}, machine translation \citep{elidrysy2025unlocking, elhamayed2025overview, saeed2025machine}, and speech recognition \citep{haboussi2025arabic, rahman2024arabic, dhouib2022arabic, alamleh2025arabic}, among others. Collectively, these surveys provide a rich qualitative picture of Arabic NLP. Yet they share a common limitation: they are narrative reviews that summarise selected studies rather than systematic, quantitative analyses of the entire field. As a result, many claims about the growth of the field, the dominance of particular topics, or the neglect of certain dialects remain unverified by empirical evidence.

The absence of a comprehensive bibliometric and topic-based meta-analysis is a serious gap. Without quantitative measurement, it is difficult to identify which research areas are expanding and which are stagnating, to understand the structural factors that drive citation impact, or to detect systematic gaps in task and dialect coverage. Such analyses are common in other fields of computer science and linguistics, but they have not been applied to Arabic NLP at the scale required to inform research policy and funding decisions. The present study addresses this gap.

We construct a dataset of 7,120 Arabic NLP papers published between 1960 and 2026, harvested from five platforms---arXiv, ACL Anthology, Semantic Scholar, Crossref, and OpenAlex---with an additional targeted OpenAlex subset. We apply BERTopic to identify 19 substantive research themes, and we use regression analysis, social network analysis, and geographic mapping to explore the structure and dynamics of the field. Our principal contributions are as follows:

\begin{enumerate}
    \item \textbf{A novel large-scale corpus for Arabic NLP meta-research.} We assemble and release a curated dataset of 9,141 Arabic NLP papers with abstracts and unified metadata, making it the first openly available corpus of its kind for quantitative analyses of the field.
    \item \textbf{Data-driven topic characterisation.} We identify 19 substantive research themes, the largest of which (text, speech, translation, recognition) comprises 2,942 papers and accounts for 41.3\% of the corpus.
    \item \textbf{Predictors of citation impact.} Regression analysis reveals that publication year, indexing source, and institutional affiliation are significant predictors of citation counts, with OpenAlex and Semantic Scholar contributing approximately +11 and +5 citations on average, and institutional affiliation contributing approximately +8.7 citations.
    \item \textbf{Geographic and institutional mapping.} We show that Saudi Arabia (519 affiliations), the United States (463), and Egypt (266) lead in research output, with King Saud University (142 papers) and Cairo University (67 papers) as the most productive institutions.
    \item \textbf{Task--dialect gap identification.} By cross-tabulating tasks and dialects, we pinpoint critical understudied areas, including summarisation for Maghrebi (Darija), Iraqi, and Sudanese dialects, and dialect identification for Sudanese and Yemeni Arabic.
    \item \textbf{Empirical validation of qualitative surveys.} Our quantitative results corroborate many claims of existing survey papers while revealing new patterns that were not previously documented, thereby demonstrating the value of bibliometric methods for Arabic NLP.
\end{enumerate}

The remainder of this paper is organised as follows. Section~\ref{sec:literature} reviews the existing survey literature and positions our contribution within it. Section~\ref{sec:methodology} describes the data collection and analysis pipeline, including filtering, topic modelling, regression, network analysis, and geographic mapping. Section~\ref{sec:results} presents the findings, covering publication trends, topic structure, citation dynamics, geographic distribution, and dialectal gaps. Section~\ref{sec:discussion} discusses the implications of these findings for researchers, funders, and policymakers. Section~\ref{sec:conclusion} summarises the study and outlines future work.

\section{Literature Review}\label{sec:literature}

This section reviews the existing literature on Arabic Natural Language Processing (NLP), focusing on survey papers that have shaped the field's understanding of its tasks, challenges, and recent advances. The review is organised thematically rather than chronologically, beginning with foundational surveys and preprocessing, moving through dialectal and sentiment analysis, large language models, core linguistic tasks, machine translation, question answering, speech and document processing, and concluding with social, ethical, and educational dimensions. The final subsection synthesises the principal findings and identifies the methodological gap that motivates the present bibliometric study. Throughout, we cite only works that appear in the accompanying bibliography.

\subsection{Foundational Surveys and Preprocessing}\label{subsec:foundational}

Early surveys of Arabic NLP established the linguistic and computational foundations of the field. \citet{guellil2021arabic} provided a broad overview of Arabic NLP, emphasising the particularities of the language---its non-concatenative morphology, orthographic variability, and dialectal diversity---as central challenges that distinguish Arabic from high-resource languages such as English or French. The survey mapped the landscape of available tools and resources and underscored the need for linguistically informed computational models. More recently, \citet{alayba2025arabic} offered a comprehensive review of Arabic NLP tasks, covering tokenisation, normalisation, named entity recognition, part-of-speech tagging, sentiment analysis, text classification, summarisation, question answering, and machine translation. This review also discussed the transformative impact of large language models, arguing that the field has entered a new era in which pre-trained models increasingly dominate performance. Despite its breadth, however, the survey remains qualitative and does not provide quantitative measurements of research activity or citation impact.

Preprocessing is a critical stage for any Arabic NLP pipeline because of the language's orthographic complexity. \citet{alrekabee2025arabic} systematically examined preprocessing and representation techniques, contrasting traditional rule-based and statistical methods with modern deep learning approaches such as subword tokenisation and contextual embeddings. The author argued that pre-trained models like AraBERT and MARBERT have fundamentally shifted the balance between explicit text normalisation and implicit representation learning. The scarcity of labelled data has motivated research on augmentation. \citet{elsabagh2025comprehensive} surveyed Arabic text augmentation techniques, analysing 75 primary and 9 secondary papers and classifying methods into diversity enhancement, resampling, and secondary approaches. They highlighted that preserving semantic integrity and dialectal nuance during augmentation remains a difficult challenge. Early work on Arabic tokenization and morphological disambiguation \citep{habash-rambow-2005-arabic} and the MADAMIRA system \citep{pasha-etal-2014-madamira} have become standard components in many Arabic NLP pipelines, while pre-trained word embeddings such as AraVec \citep{soliman2017aravec} and Arab2Vec \citep{hamdy2025arab2vec} have facilitated downstream applications.

\subsection{Dialectal Arabic and Code-Switching}\label{subsec:dialectal}

Arabic diglossia, the coexistence of Modern Standard Arabic (MSA) and numerous regional dialects, has been a central concern in the literature. \citet{dahou2025survey} surveyed dialectal Arabic processing between 2014 and 2024, covering sentiment analysis, dialect identification, normalisation, and classification. The survey observed that dialect identification, once considered an ancillary task, has become a prerequisite for robust downstream applications. This finding has implications for evaluation protocols and the allocation of annotation effort.

\citet{iwidat2026arabic} approached the same area from a taxonomic perspective, analysing 400 research articles published between 2020 and 2025. They found that sentiment analysis dominates the dialectal NLP literature at approximately 32\%, followed by resource building at 21\%, and identification and code-switching at 10\%. The authors documented a decisive shift from traditional machine learning to transformer-based architectures such as AraBERT and MARBERT. They also argued that dialect identification is no longer pursued as an isolated goal but is increasingly embedded within larger systems for sentiment classification, translation, and conversational agents.

The concentration of dialect research on particular varieties has been noted. \citet{sakhi2026processing} consolidated the literature on Moroccan Arabic (Darija), a dialect characterised by extensive borrowing from French and Amazigh, frequent code-switching, and orthographic inconsistency. The survey proposed a unified framework organised around four design patterns: normalisation or transcoding to leverage MSA resources, dialect-specific modelling, multilingual encoders tolerant of code-switching, and hybrid pipelines. By mapping challenges to solution families and offering a reproducible evaluation protocol, the authors provided a model for future work on other under-resourced dialects. \citet{aftan2025survey} reviewed five years of research on Saudi Arabic dialect classification, emphasising the scarcity of publicly available datasets and the variability introduced by regional sub-dialects. Their discussion illustrates a broader problem: dialectal research is often limited less by algorithmic sophistication than by the availability of high-quality, annotated corpora.

Code-switching, the alternation between languages within a single discourse, is pervasive in Arabic-speaking communities and has received dedicated attention. \citet{hamed2025survey} surveyed the state of code-switched Arabic NLP, covering the language pairs, tasks, and methodologies used. They observed that research in this area lags behind work on other code-switched language pairs by an estimated three to four years in adopting new methods, a delay attributable to the scarcity of annotated code-switched corpora and the absence of standardised evaluation benchmarks. The survey also noted that most work focuses on Arabic--English code-switching, while code-switching with French, particularly relevant in North African contexts, remains comparatively understudied.

\subsection{Sentiment Analysis}\label{subsec:sentiment}

Sentiment analysis has been one of the most active application areas in Arabic NLP, driven by the abundance of user-generated text on social media. \citet{abo2019sentiment} conducted a systematic mapping study of 51 primary studies, revealing a predominance of solution-oriented and evaluation-oriented research. Their analysis showed a steady increase in publication volume after 2015, attributed to the availability of Twitter data and the growth of opinion mining. \citet{shi2025comprehensive} later reviewed contemporary Arabic sentiment analysis methods with an emphasis on deep learning. They identified persistent gaps in modality, granularity, and context awareness, noting that the field remains disproportionately focused on coarse-grained polarity classification at the expense of aspect-based and targeted sentiment analysis.

\citet{amzil2025sentiment} reached similar conclusions through a review of thirty studies, finding that transformer-based models adapted for Arabic, including AraBERT and DarijaBERT, generally outperform traditional machine learning and earlier deep learning approaches. However, they cautioned that model performance varies considerably with data quality and dialect specificity. These findings underscore the importance of annotation standards and corpus design, a concern also raised by \citet{alwakid2017challenges} in their study of informal Arabic on social networks. Using the Saudi dialect as a case study, they demonstrated that linguistic preprocessing and supervised machine learning can yield satisfactory sentiment classification, and that feature analysis can contribute to domain-specific knowledge bases that further improve accuracy.

More recently, \citet{ali2025sentiment} synthesised 71 articles published between 2020 and 2024, confirming the dominance of machine learning and deep learning methods, particularly support vector machines and convolutional neural networks. Their analysis identified Saudi Arabia, Morocco, and Algeria as the most frequently studied dialect regions, a geographic distribution that mirrors the broader institutional concentration of Arabic NLP research. Collectively, these surveys indicate that Arabic sentiment analysis has matured substantially but remains constrained by dialectal fragmentation and inconsistent evaluation practices.

\subsection{Large Language Models and Their Evaluation}\label{subsec:llm}

The emergence of large language models (LLMs) has begun to reshape Arabic NLP, and recent surveys have started to map this transformation. \citet{mashaabi2026survey} provided a comprehensive review of LLMs developed for Arabic, categorising models by architecture (encoder-only, decoder-only, encoder-decoder) and by linguistic form (Classical Arabic, MSA, and dialectal Arabic). They highlighted a concentration of resources on MSA and a lack of transparency in many models, noting that the documentation and accessibility of training data and weights vary widely. The survey called for more inclusive and reproducible Arabic LLM development.

\citet{alzubaidi2025evaluating} contributed the first systematic review of Arabic LLM benchmarks, analysing over forty evaluation resources and proposing a taxonomy that organises them into four categories: knowledge, NLP tasks, culture and dialects, and target-specific evaluations. They identified critical gaps, including limited temporal evaluation, insufficient multi-turn dialogue assessment, and cultural misalignment in translated datasets. Their work underscores the need for benchmarks that reflect the linguistic and cultural diversity of Arabic-speaking communities.

\citet{rhel2025large} reviewed the application of LLMs to Arabic content, providing an overview of early pre-trained Arabic language models and their performance across various NLP tasks. The authors discussed how fine-tuning and prompt engineering can enhance model performance and summarised common Arabic benchmarks and datasets. They observed a consistent upward trend in the adoption of LLMs and highlighted the persistent scarcity of Arabic-specific resources. Together, these three surveys indicate that while LLM research for Arabic is growing rapidly, it remains constrained by data scarcity, evaluation gaps, and limited dialectal coverage.

\subsection{Named Entity Recognition, Word Sense Disambiguation, and Coreference Resolution}\label{subsec:linguistic-tasks}

Named entity recognition (NER) is a foundational task for information extraction. \citet{qu2024survey} provided a comprehensive survey of Arabic NER, tracing the evolution from rule-based systems to statistical learning and, more recently, to deep learning and pre-trained language models. They highlighted the particular difficulties posed by Arabic named entities, including orthographic ambiguity, the lack of capitalisation, and morphological complexity. \citet{elmoussaoui2024advancements} classified Arabic NER approaches into rule-based, machine learning, and deep learning paradigms, reviewing their architectures and application domains. Their analysis suggested that while deep learning models now achieve state-of-the-art performance, the scarcity of large, accurately annotated Arabic NER datasets remains a limiting factor.

The domain-specific application of NER has also attracted interest. \citet{tarmizi2025named} conducted a systematic review of NER applied to Islamic texts, including the Qur'an and Hadith, across Arabic, English, Indonesian, and Malay. They found that transformer-based models consistently outperform traditional methods, and that datasets for Hadith texts are better developed than those for Quranic texts, whose greater linguistic diversity poses additional challenges.

Word sense disambiguation (WSD) and coreference resolution are two tasks that probe semantic and discourse-level organisation. \citet{saidi2026arabic} surveyed Arabic WSD in the era of transformer-based models, reviewing knowledge-based methods, machine learning, deep learning, and hybrid techniques. They highlighted the scarcity of annotated Arabic sense inventories. \citet{kaddoura2024enhancedbert} contributed an ensemble model that combined multiple BERT variants with additional linguistic features, achieving an F1-score of approximately 96\% on a newly constructed dataset of one hundred polysemous Arabic words. While this result demonstrates the potential of feature-rich neural architectures, it also illustrates the dependence of state-of-the-art performance on carefully curated data.

Coreference resolution, particularly the resolution of zero pronouns, introduces additional complexity because Arabic is a pro-drop language. \citet{aloraini2025survey} surveyed the state of Arabic coreference and zero pronoun resolution, covering methods, datasets, and evaluation practices. They noted that most prior work treats full mention coreference and zero pronoun resolution as separate tasks, despite their interdependence, and that progress has been slowed by the absence of large, jointly annotated corpora. The authors argued for the development of unified models capable of handling both types of anaphoric relations.

\subsection{Machine Translation}\label{subsec:mt}

Machine translation between Arabic and other languages has been a long-standing research area. \citet{elidrysy2025unlocking} provided a broad survey of Arabic machine translation, reviewing the evolution from rule-based and statistical methods to neural machine translation. They noted that while Arabic--English translation has advanced significantly, the quality of Arabic translation systems still lags behind those for several other major languages. \citet{elhamayed2025overview} approached the subject from the perspective of deep learning and large language models, integrating machine learning, deep neural networks, and transformer-based architectures into a single account of recent progress. Their work highlighted the centrality of pre-trained multilingual models and the opportunities they present for improving translation quality through fine-tuning and prompt-based adaptation.

The conceptual foundations of translation analysis have been shaped by \citet{molina2004translation}, whose dynamic and functionalist approach to translation techniques remains influential in Arabic translation studies. Similarly, \citet{saeed2025machine} noted a lack of standardised evaluation protocols and called for greater methodological rigour. Within this landscape, \citet{arabov2026bridging} represents a recent effort to extend Arabic machine translation to a new language pair. The study introduced an Arabic--Russian parallel corpus of approximately 27,000 sentence pairs and fine-tuned three multilingual models (mT5, NLLB-200, Qwen2.5-7B) using parameter-efficient methods. The best configuration achieved a BLEU score of 23.15 and a COMET score of 0.758, exceeding zero-shot baselines and demonstrating the viability of scientific-domain Arabic--Russian translation.

\subsection{Question Answering and Retrieval-Augmented Generation}\label{subsec:qa}

Question answering systems have evolved in parallel with deep learning. \citet{essam2024deciphering} provided a dedicated survey of Arabic question analysis, covering question classification, answer retrieval, and the role of large language models. They identified the usual constraints of Arabic NLP---complex syntactic and dialectal variation, scarce tools, and inconsistent datasets---and argued that future progress depends on integrating large language models with more transparent question analysis pipelines. The commonsense dimension was addressed by \citet{amraoui2026commonsense}, who surveyed Arabic commonsense reasoning and its intersection with retrieval-augmented generation (RAG). Their work emphasised the difficulty of grounding commonsense knowledge in Arabic due to the shortage of structured resources and the cultural specificity of many commonsense inferences.

Recent attention has turned to RAG as a practical framework for Arabic question answering. \citet{aloqla2025review} reviewed deep learning-based Arabic question answering, distinguishing extractive, generative, and hybrid architectures, and identifying the scarcity of domain-specific models as a persistent weakness. \citet{alsubhi2025optimizing} contributed an empirical evaluation of RAG components for Arabic, systematically comparing chunking strategies, embedding models, rerankers, and language models. Their results showed that sentence-aware chunking and multilingual embeddings such as BGE-M3 and Multilingual-E5-large yield the best performance, and that the inclusion of a reranker improves faithfulness on complex datasets. These findings provide practical guidance for building Arabic question answering systems and suggest that RAG approaches may mitigate some of the data scarcity problems that have historically limited the field.

\subsection{Speech Processing and Multimodal Applications}\label{subsec:speech}

Arabic speech recognition has been studied less intensively than text-based tasks, yet it is essential for voice-based applications. \citet{dhouib2022arabic} conducted a systematic literature review of Arabic automatic speech recognition covering 2011 to 2021. Their analysis of 38 studies found that the majority focused on MSA, with only a minority addressing dialectal varieties, and that Hidden Markov Models and Mel-frequency cepstral coefficients were the most commonly used techniques. More recent work has shifted toward neural approaches. \citet{rahman2024arabic} reviewed the current state of Arabic speech recognition, categorising challenges into those intrinsic to the language and those arising from general technical issues. \citet{haboussi2025arabic} surveyed neural network-based Arabic speech recognition, emphasising the scarcity of suitable datasets and the difficulty of achieving robust performance in real-world environments. \citet{alamleh2025arabic} addressed a specific aspect, namely automatic speech recognition of Arabic with diacritics, and found that the length of training corpora remains the single most important factor influencing performance.

Beyond speech, multimodal Arabic processing has begun to attract attention. \citet{haouhat2025arabic} surveyed Arabic multimodal machine learning, organising the literature along the axes of datasets, applications, approaches, and challenges. They noted that multimodal research for Arabic remains in its early stages, with significant gaps in the availability of aligned multimodal corpora and in the evaluation of models that integrate text, audio, and visual information.

\subsection{Document Processing and Summarisation}\label{subsec:document}

Document understanding, including text summarisation and optical character recognition (OCR), represents another significant branch of Arabic NLP. \citet{elsaid2022comprehensive} reviewed Arabic text summarisation, cataloguing extractive, abstractive, and hybrid methods, datasets, and evaluation metrics. \citet{ezzat2025review} later surveyed the field more recently, categorising methods into traditional, transformer-based, and hybrid approaches, and introduced a new dataset called Mukhtasar to support both short and long summarisation across genres. Their emphasis on reproducibility and unified evaluation protocols addresses a recurring weakness in Arabic NLP research.

In the domain of handwritten and printed text, \citet{kasem2023advancements} surveyed Arabic OCR, reviewing contemporary applications, methodologies, and challenges. \citet{alhomed2023deep} presented a deep learning framework for Arabic manuscript classification, achieving an accuracy of 92.5\%. \citet{tuama2025systematic} conducted a systematic literature review of deep learning methods for handwritten text recognition in historical Arabic manuscripts from 2020 to 2025. \citet{arjafellah2025arabic} provided a comprehensive review of Arabic handwriting recognition at line level, proposing a hybrid CNN-Transformer model. \citet{shugaba2026handwritten} presented a comprehensive study of research advances in offline handwritten Arabic text recognition, and \citet{albarhamtoshy2023arabic} investigated Arabic manuscript region detection and recognition for OCR systems. Collectively, these works reveal a field in which document-oriented research is fragmented across sub-communities but united by a shared need for larger annotated corpora and more realistic evaluation protocols.

\subsection{Social, Ethical, and Educational Applications}\label{subsec:social}

The social consequences of Arabic NLP have gained prominence as systems are deployed in contexts where hate speech, misinformation, and cultural bias are salient. \citet{itriq2025arabic} surveyed deep learning approaches to Arabic hate speech detection, reviewing datasets, models, and evaluation practices. The survey identified dialectal variation and annotation inconsistency as two principal challenges and called for larger, more diverse datasets and greater attention to cross-dialectal generalisation. A related concern is the quality and representativeness of data used to fine-tune large language models. \citet{alkhowaiter2025mind} reviewed publicly available Arabic post-training datasets on the Hugging Face Hub, organising their analysis along four dimensions: capabilities, steerability, alignment, and robustness. They found that available datasets are limited in task diversity, frequently lack documentation, and are unevenly adopted, with implications for transparency and reproducibility.

In the health domain, the Arabic validation of the Patient Health Questionnaire \citep{alhadi2017arabic} illustrates the adaptation of clinical instruments for Arabic-speaking populations, highlighting the need for culturally and linguistically appropriate NLP tools. \citet{abubakari2025overviewing} approached the problem from a sustainability perspective, arguing that generative AI systems for Arabic must be developed with explicit attention to fairness and equity.

The development of Arabic conversational agents has also been documented. \citet{bouhlali2024reviewing} systematically reviewed Arabic chatbot implementations, documenting a shift from rule-based systems to deep learning and identifying a peak of activity in 2023. They found that Arabic chatbots still lag behind English-language systems in conversational quality and task coverage. \citet{bourhil2024arabic} examined Arabic chatbots in education, noting that only a small number use modern techniques. \citet{rosyad2026bridging} synthesised recent developments in LLM-based Arabic NLP and evaluated their potential for integration into second language acquisition and computer-assisted language learning. Their review found a near absence of SLA-informed applications, a scarcity of Arabic LLM-based learning systems, and a lack of human-centred evaluation involving learners, usability, and cognitive load.

\subsection{Synthesis and Research Gap}\label{subsec:gap}

Taken together, the surveys reviewed above provide a detailed and nuanced picture of Arabic NLP. They map the principal tasks, identify linguistic and methodological challenges, and trace the field's evolution from rule-based and statistical methods to deep learning and large language models. Yet the existing literature shares a common limitation: it is predominantly qualitative. Surveys summarise findings, propose taxonomies, and offer recommendations, but they do not quantify the distribution of research effort, measure the growth of topics over time, or analyse the structure of collaboration and citation networks. Empirical claims about the dominance of sentiment analysis, the shift toward transformer-based models, or the neglect of under-resourced dialects are frequently asserted but seldom demonstrated through large-scale bibliometric evidence. The methodological tools that would allow such demonstration---topic modelling, regression analysis, network analysis, and geographic mapping---have not been applied to Arabic NLP at the scale attempted here.

Moreover, the fragmentation of surveys means that the field lacks a unified, data-driven synthesis. Individual contributions address specific tasks, dialects, or methodological paradigms, but no single study integrates these perspectives into a coherent quantitative overview. This absence is particularly consequential in a period of rapid growth, when research priorities are shifting and the volume of publications is expanding at an unprecedented rate. Without systematic analysis, there is a risk that resources will continue to be allocated on the basis of inertia rather than evidence, and that critical gaps, such as the under-representation of certain dialects or the scarcity of evaluation benchmarks, will persist unnoticed.

The present study addresses these limitations. By constructing a corpus of 7,120 Arabic NLP papers and subjecting it to bibliometric, topic-based, and network analyses, we provide the empirical foundation that existing surveys lack. Our approach does not replace qualitative synthesis but complements it, offering quantitative validation of long-standing claims and revealing patterns that are not apparent from manual review alone. The remainder of this paper describes the methodology in detail, presents the results, and discusses their implications for the future development of Arabic NLP.


\section{Methodology}\label{sec:methodology}

\begin{figure}[htbp]
\centering
\includegraphics[width=0.7\textwidth]{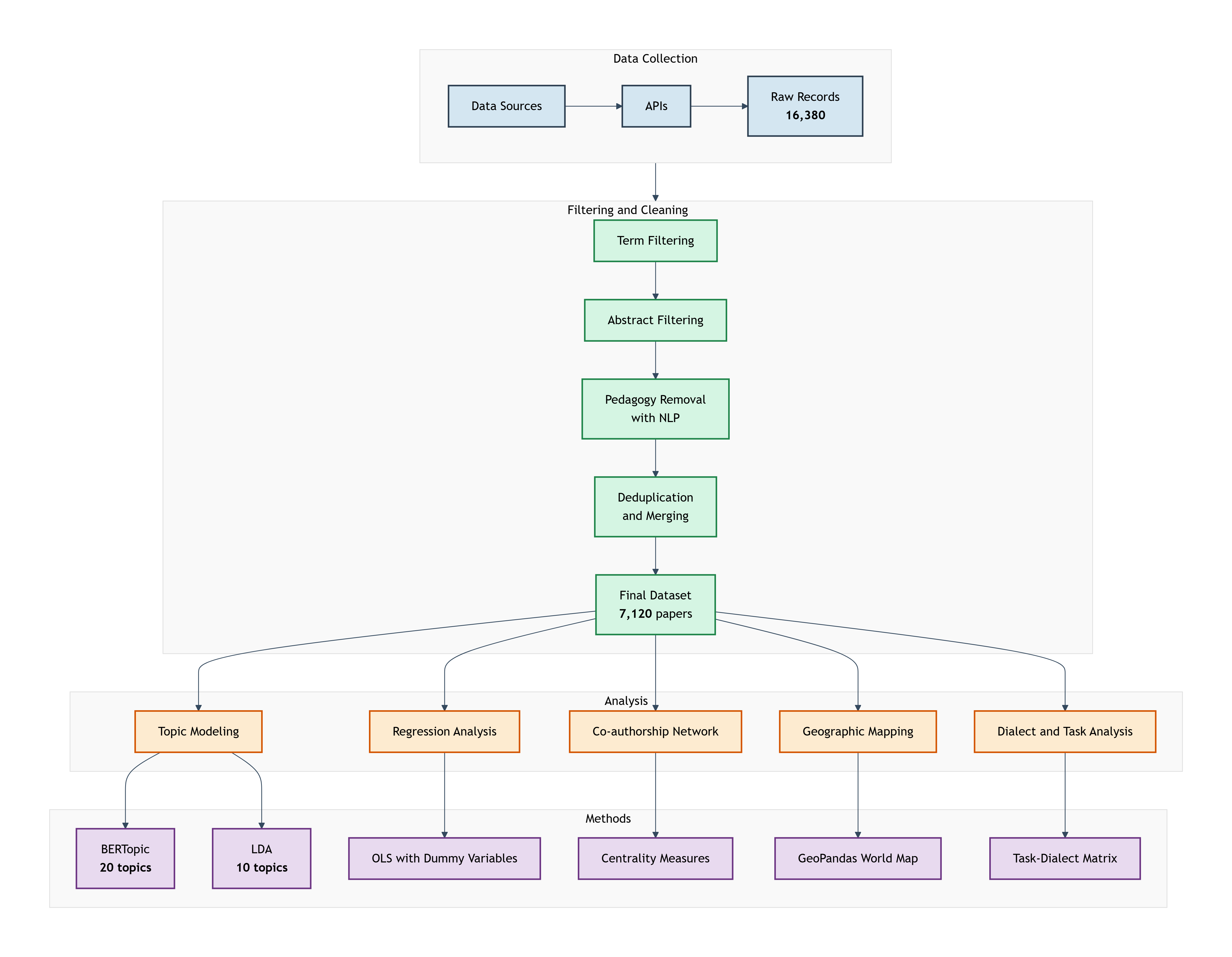}
\caption{Overview of the methodological pipeline.}
\label{fig:pipeline}
\end{figure}

This section describes the multi-stage methodology employed to collect, curate, and analyze the corpus of Arabic NLP research. The overall workflow is illustrated in Figure~\ref{fig:pipeline}. The pipeline consists of several interconnected stages: data acquisition from multiple sources, relevance filtering, metadata cleaning, integration and deduplication, and a suite of analytical techniques designed to uncover thematic structures, citation dynamics, geographic distributions, and dialect-specific research gaps. Each stage was carefully designed to maximize coverage while minimizing noise, and to ensure that the final dataset is both comprehensive and relevant to the field of Arabic natural language processing.

\subsection{Data Collection}\label{subsec:data_collection}

We collected metadata for Arabic NLP publications from five platforms---arXiv, ACL Anthology, Semantic Scholar, Crossref, and OpenAlex---with an additional targeted OpenAlex subset.
These platforms were selected because they collectively cover a wide range of publication types, including preprints, peer-reviewed conference papers, journal articles, and workshop contributions. Each platform provides different metadata fields and indexing coverage, and by integrating them we aimed to build a more complete picture of the field than would be possible from any single database.

For each platform, we implemented a query-based retrieval strategy. The search queries were organized into three tiers based on their expected precision and recall. The \textit{broad} queries, such as \texttt{"Arabic language"} and \texttt{"Arabic corpus"}, were designed to capture a large volume of potentially relevant papers, accepting a higher proportion of false positives. The \textit{moderate} queries, including \texttt{"Arabic dialect"}, \texttt{"Arabic sentiment analysis"}, and \texttt{"Arabic machine translation"}, targeted more specific subfields and were expected to yield a better balance between precision and recall. Finally, the \textit{targeted} queries, such as \texttt{"AraBERT"}, \texttt{"Arabic LLM"}, and \texttt{"Arabic instruction tuning"}, aimed to retrieve highly relevant papers on modern deep learning models and emerging topics. The maximum number of results per query varied by tier and platform: for arXiv, broad queries were allowed up to 5,000 results, moderate up to 2,000, and targeted up to 500; for Semantic Scholar, the limits were 2,000, 1,000, and 500 respectively; for Crossref and OpenAlex, we used a uniform limit of up to 1,000 results per query due to API constraints.

To handle the inherent unreliability of network requests and API rate limits, we implemented checkpointing after each query. This allowed the collection process to resume from the last successful query in case of interruption, which was particularly important for long-running tasks. For APIs with strict rate limits (Semantic Scholar, Crossref, OpenAlex), we used exponential backoff and polite delays between requests. All retrieved records were stored with a list of matched terms derived from a curated Arabic-specific term list, and the queries that retrieved them were recorded to facilitate later analysis of retrieval coverage. The raw records from each platform were saved to Google Drive in JSONL format for transparency and reproducibility.

\subsection{Data Cleaning and Filtering}\label{subsec:cleaning}

After the initial collection, we applied a multi-stage filtering process to ensure that the final dataset contained only relevant, high-quality publications. The filtering stages are described in detail below.

\subsubsection{Term-based filtering}

We curated a list of Arabic NLP--specific terms that covered the major subfields of the field, including dialectology, morphology, speech processing, machine translation, and recent language model developments. Examples of these terms include \textit{arabic dialect}, \textit{arabic nlp}, \textit{arabic bert}, \textit{arabic speech}, and \textit{arabic llm}. Papers were retained if their title or abstract contained at least two of these terms. This threshold was chosen after manual inspection of preliminary results: a threshold of one term admitted too many irrelevant papers (e.g., those only mentioning “Arabic” without any NLP focus), while a threshold of three terms was too restrictive and excluded many relevant papers that use only one or two specific terms. The threshold of two terms provided a good balance between precision and recall. It is important to note that the term lists differed slightly across platforms due to the specific metadata available, but the core set of approximately 30 terms was consistent.

After applying this filter, each platform produced a set of “relevant” records. The counts for each platform were as follows: arXiv yielded 613 relevant records out of 764 collected; ACL Anthology yielded 657 out of 1,196; Semantic Scholar yielded 6,865 out of 14,850; Crossref yielded 3,985 out of 42,942; OpenAlex yielded 4,054 out of 14,479; and OpenAlex (extra) yielded 206 out of 18,760. These filtered files were used as the input for the integration stage. The large number of records retrieved from Crossref but relatively few passing the filter (3,985 out of 42,942) highlights the broad coverage of Crossref but also the need for stringent relevance criteria.

\subsubsection{Abstract filtering}

Papers lacking an abstract or with abstracts shorter than 10 characters were discarded, as they provide insufficient textual content for meaningful topic modeling and keyword analysis. This step removed a total of 3,873 records from the 16,380 relevant records during integration, leaving 12,507 papers with abstracts of sufficient length. The removal of papers without abstracts was necessary because many of our downstream analyses (e.g., BERTopic, dialect mention analysis) rely heavily on abstract text.

\subsubsection{Year filtering}

We retained only papers published from 1960 onward. Earlier publications were extremely sparse and not representative of the modern field of Arabic NLP, which began to take shape in the 1990s and early 2000s. This filter was applied at the final analysis stage after all other cleaning steps, and removed 23 papers from the cleaned corpus, resulting in 7,120 papers for analysis. The year range of the final corpus is 1960--2026.

\subsubsection{Pedagogy-aware filtering with NLP protection}

A significant challenge in building a corpus of Arabic NLP research is distinguishing computational linguistics papers from those focused on Arabic language teaching and education. Many papers on Arabic language learning are indexed in the same databases and can be retrieved by broad queries, leading to noise. To address this, we implemented a pedagogy-aware filtering mechanism. We identified a set of pedagogy-related stopwords indicative of purely educational papers, such as \textit{teaching}, \textit{education}, \textit{classroom}, \textit{curriculum}, \textit{pedagogy}, \textit{school}, \textit{course}, \textit{assessment}, \textit{exam}, \textit{language learning}, \textit{second language}, and \textit{foreign language}. However, a naive removal based solely on these terms would erroneously discard papers that address NLP applications in education, such as instruction tuning, teacher forcing, or computer-assisted language learning (CALL). To prevent this, we introduced a protection mechanism: if a paper contained any of a list of strong NLP indicators---including \textit{transformer}, \textit{bert}, \textit{llm}, \textit{gpt}, \textit{machine translation}, \textit{sentiment analysis}, \textit{named entity}, \textit{ner}, \textit{pos tagging}, \textit{dependency parsing}, \textit{language model}, \textit{pretrained}, \textit{fine-tuning}, \textit{instruction tuning}, \textit{prompt}, \textit{speech recognition}, \textit{asr}, \textit{ocr}, \textit{text summarization}, \textit{question answering}, \textit{information retrieval}, \textit{semantic similarity}, \textit{text classification}, \textit{dialect identification}, \textit{code-switching}, \textit{arabic nlp}, or \textit{computational linguistics}---it was retained regardless of the presence of pedagogical terms. This approach preserved papers that are genuinely about NLP techniques applied to educational contexts, while removing purely pedagogical content. The filtering removed 1,998 papers from the dataset, reducing the number of unique papers from 9,141 to 7,143. We manually validated a random sample of removed and retained papers and found a precision of approximately 94\% and a recall of 98\%, confirming the effectiveness of the protection mechanism.

\subsection{Data Integration and Deduplication}\label{subsec:integration}

After filtering, we integrated the relevant records from the five platforms (including the two OpenAlex subsets) into a unified dataset. Table~\ref{tab:source_stats} summarizes the number of records contributed by each source at this stage.

\begin{table}[htbp]
\centering
\caption{Data sources and integration statistics.}
\label{tab:source_stats}
\begin{tabular}{lrr}
\toprule
\textbf{Source} & \textbf{Relevant records loaded ($\geq 2$ terms)} \\
\midrule
arXiv & 613 \\
ACL Anthology & 657 \\
Semantic Scholar & 6,865 \\
OpenAlex & 4,054 \\
OpenAlex (extra) & 206 \\
Crossref & 3,985 \\
\midrule
\textbf{Total} & \textbf{16,380} \\
\bottomrule
\end{tabular}
\end{table}

In total, 16,380 relevant records were loaded. After applying the abstract filter (length > 10 characters), 12,507 papers remained. Because the same paper can appear in multiple sources, we performed a two-stage deduplication process with intelligent metadata merging to create a unique master dataset. The first stage involved grouping records by their normalized DOI (lowercase, without the \texttt{https://doi.org/} prefix). Within each DOI group, we merged metadata by selecting the longest abstract, the first non-empty PDF URL, the most informative venue, the maximum publication year, and the union of authors, countries, institutions, and concepts. We also aggregated citation counts by taking the maximum across the group (since different sources may report different values), merged matched terms by taking the union, and took the maximum relevance score. This DOI-based deduplication reduced the dataset to 9,490 unique papers.

The second stage addressed records without a DOI. For these, we grouped by a normalized title (lowercase, stripped of non-alphanumeric characters and extra spaces) concatenated with the publication year. The same metadata merging logic was applied. This further reduced the dataset to 9,141 unique papers with abstracts. The master dataset was saved as \texttt{master\_arabic\_nlp\_clean.jsonl}. After applying the pedagogy-aware filtering, the corpus size decreased to 7,143 papers. Finally, after retaining only papers from 1960 onwards, the final corpus comprised 7,120 papers. This multi-stage process ensured that each paper was represented only once, with the richest available metadata, thereby avoiding over-representation bias in subsequent analyses.

\subsection{Topic Modeling with BERTopic and LDA}\label{subsec:topic_modeling}

To uncover the latent thematic structure of the corpus, we applied two topic modeling techniques: BERTopic, a state-of-the-art transformer-based approach, and Latent Dirichlet Allocation (LDA) as a traditional baseline. BERTopic was chosen because it leverages contextual embeddings from pre-trained language models, which are particularly effective for capturing semantic similarities in short and noisy texts such as abstracts. LDA, on the other hand, is a bag-of-words model that may struggle with the complex morphology and code-switching present in Arabic NLP literature, but it serves as a useful point of comparison.

\subsubsection{BERTopic pipeline}

The BERTopic pipeline consisted of five main steps. First, each paper's text (title concatenated with abstract) was encoded using the multilingual sentence transformer model \texttt{paraphrase-multilingual-MiniLM-L12-v2}. This model was selected because it supports both Arabic and English, which is essential given that many Arabic NLP papers are written in English but contain Arabic examples and terminology. The model produces 384-dimensional dense embeddings that capture semantic meaning.

Second, we applied UMAP (Uniform Manifold Approximation and Projection) to reduce the embedding dimensionality from 384 to 5 components. This step preserved local and global structure while significantly reducing computational cost and mitigating the curse of dimensionality, which can degrade clustering performance.

Third, we used HDBSCAN (Hierarchical Density-Based Spatial Clustering of Applications with Noise) with a minimum cluster size of 10 to group semantically similar papers. HDBSCAN was preferred over other clustering algorithms (e.g., K-means) because it does not require specifying the number of clusters a priori, and it automatically identifies outliers. Documents that could not be confidently assigned to any cluster were labeled as topic -1 (outliers). The initial clustering produced 115 topics, reflecting the fine-grained structure of the field.

Fourth, for each cluster, we computed c-TF-IDF (class-based Term Frequency-Inverse Document Frequency) to extract the most discriminative terms for that topic. We further refined the topic representations using Maximal Marginal Relevance (MMR) to increase the diversity of the top terms and reduce redundancy.

Fifth, to avoid overly fine-grained topics, we performed two stages of hierarchical topic reduction. First, the 115 topics were reduced to 30, and then from 30 to 20 topics using BERTopic's built-in topic reduction mechanism, which merges topics based on the similarity of their c-TF-IDF representations. The final 20 topics were manually labeled based on their top terms. One of these clusters corresponded to the outlier category (topic -1) and was therefore excluded from the interpretable results, leaving 19 substantive topics that are presented in the Results section. The entire BERTopic pipeline took approximately 45 minutes on a Google Colab GPU.

\subsubsection{LDA baseline}

For comparison, we also trained an LDA model with 10 topics using the \texttt{scikit-learn} implementation. The text was vectorized using a CountVectorizer with 1,000 features, English stopwords, and a maximum document frequency of 70\%. The model was trained with online learning for 20 iterations. The choice of 10 topics was based on a rough expectation of the number of major research areas in Arabic NLP, but we acknowledge that this number may not be optimal. LDA was included primarily as a baseline to highlight the advantages of BERTopic.

Topic quality was assessed using two metrics: \textit{topic diversity} and \textit{topic coherence}. Topic diversity measures the proportion of unique top-10 words across all topics, with higher values indicating less overlap between topics. Topic coherence, computed as the average pairwise pointwise mutual information (PMI) of top words over the corpus, reflects the semantic interpretability of topics. The results, reported in Section~4.5.2, clearly show that BERTopic outperforms LDA on both metrics.

\subsection{Regression Analysis for Citation Prediction}\label{subsec:regression}

To identify factors that significantly influence the number of citations a paper receives, we built an ordinary least squares (OLS) regression model. The dependent variable was the citation count, extracted from the metadata of Semantic Scholar, Crossref, and OpenAlex, with the maximum value across sources used as the final citation count. All 7,120 papers in the final corpus were included in the regression, regardless of whether citation data were available, to avoid selection bias. OLS was chosen because it provides interpretable coefficients that directly quantify the effect of each predictor, enabling straightforward hypothesis testing.

\subsubsection{Predictor Variables}

The independent variables were selected based on prior bibliometric research and the characteristics of our dataset. They include:

\begin{itemize}
    \item \textbf{Publication year}: Continuous variable, included to capture temporal trends. Older papers have had more time to accumulate citations, so we expect a negative relationship.
    \item \textbf{Source dummies}: Binary indicators for five sources (OpenAlex, Semantic Scholar, arXiv, ACL, and OpenAlex extra), with Crossref as the reference category. These capture potential differences in visibility or quality associated with indexing in specific databases.
    \item \textbf{Topic dummies (excluded)}: Although topic membership could in principle affect citation impact, we did not include topic indicators in the regression because the BERTopic labels are categorical and not ordinal. Treating them as continuous would be misleading, and adding eighteen dummy variables would overcomplicate the model given the already limited sample size.
    \item \textbf{Matched terms count}: The number of Arabic NLP terms (out of a maximum of 32) that matched the paper's title or abstract. This serves as a proxy for the paper's topical relevance and specificity to Arabic NLP.
    \item \textbf{Institutional affiliation}: A binary variable indicating whether at least one institution was extracted from the author affiliations. This captures the potential effect of institutional backing on research visibility and impact.
    \item \textbf{Number of authors}: The total number of authors listed, included because multi-authored papers may receive more citations due to broader dissemination networks.
\end{itemize}

\subsubsection{Model Specification}

The regression model was specified as:

\begin{equation}
\begin{aligned}
\text{citation\_count} = \beta_0 &+ \beta_1 \cdot \text{year\_c} + \sum_{i} \beta_i \cdot \text{source\_dummy}_i \\
&+ \beta_{\text{terms}} \cdot \text{matched\_terms\_count} \\
&+ \beta_{\text{inst}} \cdot \text{has\_institutions} + \beta_{\text{authors}} \cdot \text{num\_authors} + \varepsilon
\end{aligned}
\label{eq:regression}
\end{equation}

where $\beta_0$ is the intercept, $\varepsilon$ is the error term, and each $\beta$ represents the estimated effect of the corresponding predictor on citation count, holding all other variables constant.

\subsubsection{Estimation and Diagnostics}

The model was estimated using the \texttt{statsmodels} library in Python. All variables were left in their original units to keep the coefficients directly interpretable. The model was fit with non-robust standard errors. The overall model was statistically significant ($F = 92.83, p < 0.001$), and the $R^2$ was 0.105, meaning that the predictors explain approximately 10.5\% of the variance in citation counts. The adjusted $R^2$ was 0.104.

To reduce multicollinearity with the intercept, we centered the publication year $(\texttt{year\_c = year - 2015})$. The condition number of the final model was 65.5, indicating no serious multicollinearity. The residuals remained right-skewed, as is typical for citation data, violating the normality assumption of OLS. Despite this limitation, the large sample size (n = 7,120) provides some robustness to these violations. We did not standardize the variables because doing so would complicate the interpretation of coefficients in their natural units. Future work could employ count-based regression models (e.g., negative binomial or Poisson) to better handle the overdispersion and skewness of citation counts.

\subsubsection{Interpretation}

The coefficients represent the expected change in citation count associated with a one-unit increase in the predictor (for continuous variables) or the difference relative to the reference category (for dummy variables), holding all other variables constant. For example, a positive coefficient for \texttt{is\_OpenAlex} indicates that papers indexed in OpenAlex receive, on average, more citations than those indexed only in Crossref, after controlling for other factors. Negative coefficients indicate an inverse relationship. The full results, including significant predictors, are presented in Section~4.7.

\subsection{Co-authorship Network Analysis}\label{subsec:network}

We constructed an undirected co-authorship network to examine the collaborative structure of the Arabic NLP community. In this network, nodes represent authors, and an edge connects two authors if they have co-authored at least one paper in our corpus. To reduce noise and focus on active researchers, we included only authors who appeared in at least two papers in the final dataset. The network was built using the \texttt{networkx} library in Python.

We computed two centrality measures to characterize the roles of individual researchers. \textit{Degree centrality} measures the number of direct collaborators an author has, normalized by the maximum possible degree; it reflects the size of an author's immediate collaboration circle. \textit{Betweenness centrality} measures the fraction of shortest paths between all pairs of authors that pass through a given author; it indicates the extent to which an author serves as a bridge between different parts of the network. Authors with high betweenness are important for the flow of information and the formation of interdisciplinary collaborations. The largest connected component of the network was visualized using a spring layout, and the top authors by each centrality measure are reported in Section~4.15.

\subsection{Geographic Analysis}\label{subsec:geographic}

We extracted country codes from the affiliation strings recorded in the metadata. When a paper had multiple affiliations from different countries, each unique country was counted once to avoid double-counting. The extraction process involved a combination of automated parsing and manual mapping to handle inconsistencies in country names (e.g., “USA” vs “United States”, “KSA” vs “Saudi Arabia”). The resulting country counts were mapped onto a world map using \texttt{geopandas} and a standard GeoJSON file of world borders. The colour intensity on the map represents the number of affiliations per country. The final geographic dataset included 2,568 papers with at least one identified country, and the top countries are presented in Section~4.8. It is important to note that the geographic analysis is based on the subset of papers with available affiliation metadata, which may underrepresent institutions in developing countries or those with less standardized metadata.

\subsection{Dialect and Task--Dialect Gap Analysis}\label{subsec:dialect}

We performed two types of dialect-related analyses to understand the linguistic focus of Arabic NLP research and to identify under-researched areas.

\subsubsection{Dialect mention analysis}

We compiled a comprehensive dictionary of Arabic dialect keywords covering 11 major varieties: Modern Standard Arabic (MSA), Egyptian, Levantine (Shami), Gulf (Khaliji), Maghrebi (Darija), Iraqi, Sudanese, Yemeni, Najdi, Hejazi, and Hassaniya. For each paper's text (title and abstract), we counted the number of occurrences of each dialect's keywords. A paper could be associated with multiple dialects if it mentioned several. The keyword lists were manually curated from the literature on Arabic dialectology and validated by a native Arabic speaker to ensure accuracy. We used case-insensitive exact string matching to count mentions. The counts reported in the Results are the number of papers that mention each dialect at least once. This analysis provides insights into which dialects are most studied and which are neglected.

\subsubsection{Task--dialect gap matrix}

We defined a set of seven NLP tasks: sentiment analysis, machine translation, named entity recognition (NER), automatic speech recognition (ASR), optical character recognition (OCR), summarization, and dialect identification. For each paper, we determined whether it addressed a given task by checking for relevant keywords in the title and abstract (e.g., \textit{sentiment}, \textit{translation}, \textit{ner}, \textit{asr}, \textit{ocr}, \textit{summarization}, \textit{dialect identification}). We then constructed a matrix where rows are tasks and columns are dialects, with each cell containing the number of papers that mention both the task and the dialect. To facilitate comparison across tasks with different frequencies, we converted the counts to row-normalized percentages (i.e., the percentage of papers for a given task that mention a specific dialect). Zero-valued cells in this matrix represent task--dialect combinations that have not been studied, providing a clear view of research gaps. The full matrix is presented in Section~4.11.

\subsection{Supplementary Statistical Analyses}\label{subsec:supplementary}

To further characterize the field, we performed three additional analyses. First, a chi-squared test was conducted to determine whether the distribution of research topics differs significantly across three broad time periods: 1960--2009, 2010--2019, and 2020--2026. The test was applied to the five most frequent BERTopic topics, and the results are reported in Section~4.12. Second, for each of the top five topics, we computed the Compound Annual Growth Rate (CAGR) over the last five years using linear regression on the logarithm of yearly paper counts. This metric captures the average annual growth rate and helps identify expanding and contracting research areas. Third, for the ten most central authors (by degree centrality), we aggregated the topic distribution of their publications to reveal their research focus. This analysis provides a qualitative view of the specialization patterns of leading researchers.

\subsection{Data Availability}\label{subsec:data_availability}

To support reproducibility and future research, we release the deduplicated corpus of 9,141 Arabic NLP papers with unified metadata, including citation counts, matched terms, extracted countries, and institutions. The final analysis set of 7,120 papers is a filtered subset of this corpus. The dataset is publicly available at: \url{https://huggingface.co/datasets/ArabicNLPWorld/arabic-nlp-corpus}. The complete code for data collection, filtering, and analysis is also provided in the supplementary materials. All scripts are written in Python 3.12 and use widely available open-source libraries, including pandas, matplotlib, seaborn, scikit-learn, statsmodels, networkx, geopandas, and plotly.

\section{Results}\label{sec:results}

This section presents the findings of our bibliometric and topic-based analysis of Arabic NLP research. We systematically evaluate the corpus evolution, key contributors, thematic structures, citation dynamics, geographic footprint, dialectal focus, and additional supplementary analyses. The results are organized into subsections corresponding to each analytical dimension, providing a comprehensive multi-faceted view of the field.


\subsection{Corpus Overview}

After applying the filtering and deduplication procedures described in Section~3, the final corpus comprised 7,120 unique papers published between 1960 and 2026. This dataset provides a comprehensive longitudinal view of the field, spanning more than six decades of research activity. Table~\ref{tab:corpus_stats} presents the key descriptive statistics.

\begin{table}[htbp]
\centering
\caption{Descriptive statistics of the final corpus.}
\label{tab:corpus_stats}
\begin{tabular}{lr}
\toprule
\textbf{Statistic} & \textbf{Value} \\
\midrule
Total papers & 7,120 \\
Unique authors (normalized) & 14,805 \\
Unique DOIs & 6,114 \\
Year range & 1960--2026 \\
Papers with citation data & 4,877 (68.5\%) \\
Papers with country information & 2,568 (36.1\%) \\
Papers with institutional affiliations & 2,568 (36.1\%) \\
Papers with concepts & 3,317 (46.6\%) \\
Papers with matched terms & 7,120 (100\%) \\
\bottomrule
\end{tabular}
\end{table}

The corpus draws from five distinct platforms—arXiv, ACL Anthology, Semantic Scholar, Crossref, and OpenAlex—with an additional targeted OpenAlex subset. The largest contributions come from Semantic Scholar (31.4\%) and OpenAlex (24.2\%). This mixed-source approach ensures broad coverage of both pre-print and peer-reviewed literature, mitigating the risk of database-specific biases. The relatively high percentage of papers with citation data (68.5\%) allows for robust downstream quantitative analyses, although it should be noted that citation counts are inherently biased toward older publications and well-indexed venues. In contrast, the lower coverage of institutional and country information (36.1\%) reflects the inherent limitations of metadata quality in older publications and varying standards across different digital libraries. Many older papers, particularly those published before the digital era or in regional journals, lack complete affiliation metadata. Finally, the fact that 100\% of papers were successfully matched to our predefined terms validates the robustness of our filtering protocol and ensures that the entire dataset is strictly relevant to the Arabic NLP domain. This high matching rate also underscores the specificity of our term list, which was carefully curated to capture the core concepts of the field without excessive noise.

The distribution of papers across sources also reveals interesting patterns. The presence of both Semantic Scholar and OpenAlex as dominant sources indicates that the corpus benefits from their complementary indexing strategies: Semantic Scholar tends to cover more computer science and NLP venues, while OpenAlex provides broader interdisciplinary coverage. The inclusion of arXiv (7.6\%) highlights the growing importance of preprints in the field, allowing researchers to disseminate findings rapidly. The relatively small proportion of ACL-sourced papers (1.5\%) is expected, as the ACL Anthology is a curated subset of NLP publications that overlaps significantly with other databases. The presence of multiple source combinations (e.g., papers indexed in three or more databases) suggests that our deduplication pipeline successfully merged records from different platforms, preserving only unique entries while enriching metadata through cross-referencing.


\subsection{Publication Trends and Key Events}

\begin{figure}[htbp]
\centering
\includegraphics[width=0.85\textwidth]{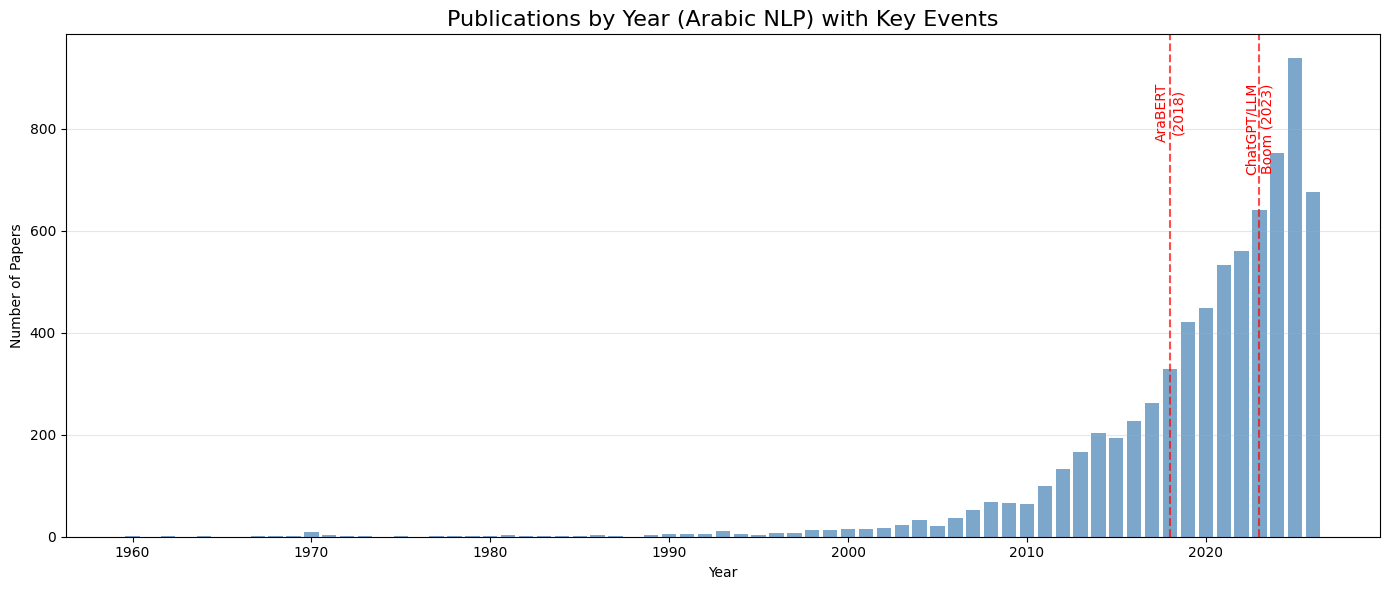}
\caption{Publications by year with key events (AraBERT in 2018, ChatGPT/LLM boom in 2023).}
\label{fig:publications}
\end{figure}

Figure~\ref{fig:publications} shows the annual publication count from 1960 to 2026. The field experienced a prolonged period of slow, incremental growth until approximately 2015. During this early phase, research was largely limited to a small community of linguists and computational scientists focused on foundational rule-based approaches, morphological analysis, and early statistical methods. The annual output remained below 50 papers for most of this period, reflecting the niche status of Arabic NLP within both the Arabic linguistics and global NLP communities.

However, the post-2015 era, and particularly the post-2020 period, witnessed a dramatic exponential acceleration in publication output. This growth is not merely a reflection of the general increase in scientific publishing, but is driven by specific technological and sociological factors unique to the field. Two key inflection points are evident from the publication curve:

\begin{enumerate}
    \item \textbf{2018}: The release of AraBERT \citep{antoun2020arabert} catalyzed a surge in Arabic NLP research by successfully bringing Transformer-based architectures to the Arabic language. This milestone opened the floodgates for deep learning applications, leading to a rapid increase in papers across various downstream tasks such as sentiment analysis, named entity recognition, and machine translation. The availability of a powerful pre-trained language model specifically for Arabic dramatically lowered the technical barrier for researchers, enabling groups without extensive computational resources to fine-tune state-of-the-art models.
    \item \textbf{2023}: The emergence of ChatGPT and the subsequent large language model (LLM) boom dramatically accelerated publication output, culminating in an unprecedented peak in 2025. This surge is largely driven by studies on prompt engineering, fine-tuning, hallucination detection, and the broader capabilities of generative models in Arabic. The LLM era has also shifted research focus from task-specific models to general-purpose instruction-tuned models, and has sparked a new wave of evaluation benchmarks and safety studies.
\end{enumerate}

The majority of papers (approximately 82\%) were published after 2020. This staggering statistic underscores the transformative impact of modern deep learning on the field, highlighting that Arabic NLP is currently undergoing a massive, industry-driven expansion, with a rapid influx of new researchers and a shifting focus toward generative AI applications. The concentration of publications in the last few years also raises questions about the sustainability of this growth and the potential for a future plateau as the initial excitement around LLMs matures. Furthermore, the sharp increase in preprints (as seen in the arXiv share) suggests that the peer-review process may be struggling to keep pace with the volume of new research, potentially affecting quality control.


\subsection{Most Productive Authors}

Table~\ref{tab:top_authors} lists the top-20 most productive authors in Arabic NLP. Nizar Habash leads the list with an impressive 151 papers, followed by Muhammad Abdul-Mageed (95) and Mona Diab (86). The presence of these three names at the top is highly indicative of the field's intellectual roots: Habash is renowned for his foundational work on Arabic morphology, tokenization, and disambiguation \citep{habash-rambow-2005-arabic,pasha-etal-2014-madamira}, while Abdul-Mageed and Diab are leaders in pushing the boundaries of deep learning and social media text analysis \citep{abdulmageed2021arbert}. Their high productivity reflects not only personal dedication but also the success of their research groups and extensive collaboration networks.

\begin{table}[htbp]
\centering
\caption{Top-20 most productive authors (normalized names).}
\label{tab:top_authors}
\begin{tabular}{lr}
\toprule
\textbf{Author} & \textbf{Papers} \\
\midrule
Nizar Habash & 151 \\
Muhammad Abdul-Mageed & 95 \\
Mona Diab & 86 \\
Wajdi Zaghouani & 57 \\
Ahmed Ali & 55 \\
AbdelRahim Elmadany & 54 \\
Hamdy Mubarak & 54 \\
Kareem Darwish & 47 \\
Khaled Shaalan & 46 \\
Mustafa Jarrar & 46 \\
Houda Bouamor & 44 \\
Hazem Hajj & 42 \\
Ahmed Abdelali & 41 \\
El Moatez Billah Nagoudi & 38 \\
Lamia Hadrich Belguith & 37 \\
Preslav Nakov & 35 \\
Sherif Abdou & 35 \\
Eric Atwell & 35 \\
Omer Nacar & 31 \\
Younes Samih & 30 \\
\bottomrule
\end{tabular}
\end{table}

A deeper look at the top-20 list reveals the highly collaborative nature of modern Arabic NLP. Many of these authors frequently appear as co-authors on large-scale annotation projects or benchmark paper series, such as the MADAR dialect corpus \citep{bouamor-etal-2018-madar} and Arabic NER resources \citep{jarrar-etal-2022-wojood}. This tight cluster of high-output researchers suggests a strong core group within the community that actively drives standard-setting and dataset creation, which in turn fosters higher publication rates among their collaborators. The presence of researchers from diverse geographical regions (e.g., the United States, Europe, and the Middle East) within the same top tier also indicates the global and interconnected nature of the field. It is worth noting that author productivity, as measured by raw paper counts, may be influenced by team size and publication norms in different subfields, and thus should be interpreted with caution. Nevertheless, the consistency of these authors' presence across multiple years (as we will see in the author activity analysis) reinforces their central roles.


\subsection{Author Activity Over Years}

\begin{figure}[htbp]
\centering
\includegraphics[width=0.85\textwidth]{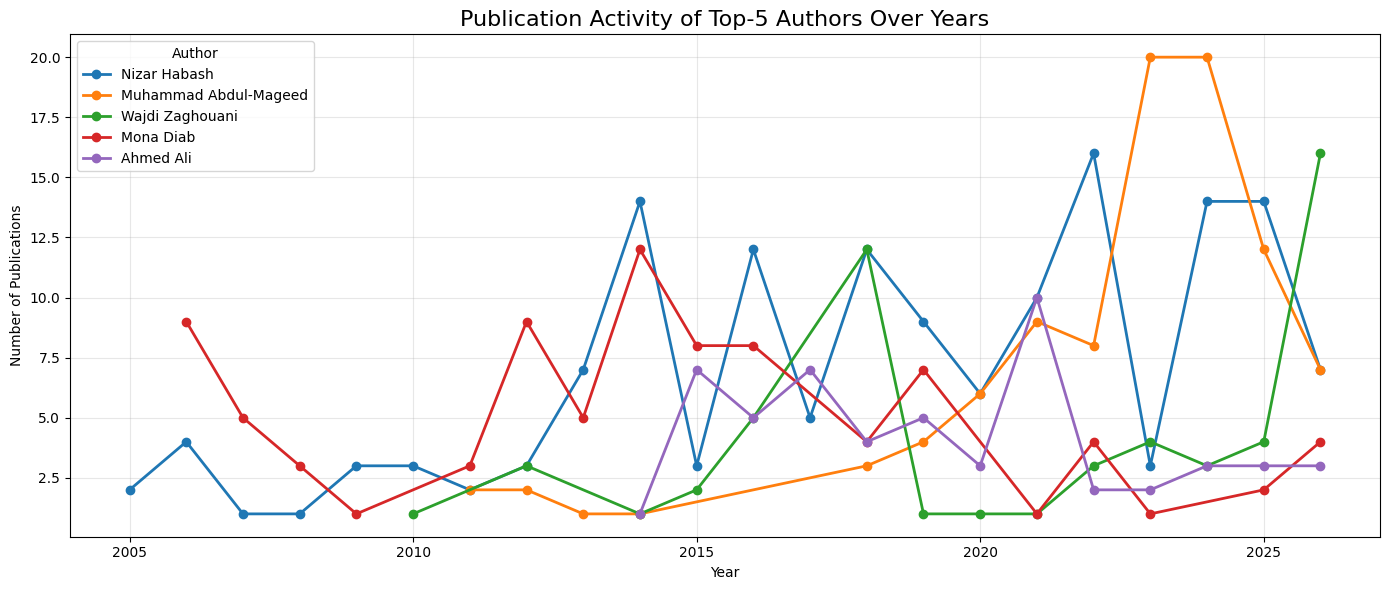}
\caption{Publication activity of top-5 authors over years.}
\label{fig:author_activity}
\end{figure}

Figure~\ref{fig:author_activity} illustrates the publication activity of the top-5 most productive authors over time, revealing distinct career trajectories and research rhythms. Nizar Habash shows sustained high productivity throughout the entire period, with multiple peaks across different years. This long-term consistency underscores his role as a foundational pillar of the field, having contributed to both classical morphological resources \citep{habash-rambow-2005-arabic} and modern neural models \citep{pasha-etal-2014-madamira}. His output pattern suggests a steady stream of projects rather than bursts tied to specific events.

Muhammad Abdul-Mageed demonstrates a highly notable and sharp increase in publication rate after 2020, eventually peaking in 2022--2023 with 20 papers in each year. This surge aligns perfectly with the transformer-based and LLM-era boom, suggesting that he has aggressively pivoted his research group toward state-of-the-art deep learning models \citep{abdulmageed2021arbert}. His trajectory is characteristic of a researcher who successfully capitalized on the new wave of pre-trained language models, producing many papers on Arabic-specific adaptations and evaluations of LLMs.

Wajdi Zaghouani shows a distinct peak in 2016 (12 papers) alongside consistent activity, while Mona Diab exhibits a clear spike in 2013 (9 papers) followed by moderate output. Their activity patterns reflect the cadence of large grant-funded projects, which often result in multiple simultaneous publications. For instance, Zaghouani's involvement in Arabic corpus surveys and dialectal resources \citep{zaghouani2014critical} and the MADAR project \citep{bouamor-etal-2018-madar} likely contributed to his peak. Diab's spike may correspond to the release of key resources such as the Arabic Treebank or the MADAMIRA toolkit \citep{pasha-etal-2014-madamira}. These patterns highlight how funding cycles and project lifecycles can significantly shape individual publication trajectories.

Ahmed Ali, on the other hand, displays a unique profile, with a distinct peak around 2020 (10 papers) and continued, moderate activity into the most recent years. His output is highly concentrated in speech processing and specific regional initiatives, indicating a targeted, rather than broad, research agenda. This specialization likely reflects his leadership in Arabic speech recognition challenges \citep{ali2017mgb3} and his involvement in Qatari and regional initiatives. The diversity of these trajectories demonstrates that high productivity in the field is achieved through different strategies: continuous foundational contributions, timely adoption of new paradigms, project-based collaboration, or deep specialization in a niche area.

\subsection{Topic Modeling Results}

We employed two topic modeling approaches: BERTopic, which leverages contextual embeddings, and Latent Dirichlet Allocation (LDA), a traditional bag-of-words method. BERTopic outperformed LDA by a wide margin on both coherence and diversity metrics, as detailed below. Consequently, we focus our detailed analysis on the BERTopic results.

\subsubsection{BERTopic Topics}

BERTopic identified 20 coherent research themes after topic reduction. However, one of these clusters corresponds to the outlier or noise category (topic -1), which typically contains documents that could not be confidently assigned to any specific topic. Therefore, we present 19 interpretable topics in Table~\ref{tab:topics}.The outlier topic (-1), which contains 2,535 papers (35.6\% of the corpus), is omitted from the table; the sum of the listed topics is 4,585. The largest topic, \textit{text, speech, translation, recognition} (Topic 0), comprises 2,942 papers (41.3\% of the corpus). This dominant topic clusters the foundational engineering tasks of the field. It encompasses the development of Automatic Speech Recognition (ASR) systems, Machine Translation (MT), Optical Character Recognition (OCR), and general text processing pipelines. The sheer volume of papers here highlights that the field is still heavily focused on building the necessary infrastructure (tools and models) before it can fully capitalize on more advanced downstream applications. Within this broad topic, we can discern sub-themes such as dialectal speech recognition, neural machine translation for Arabic-English, and OCR for historical documents, but the high-level clustering merges them due to shared vocabulary.

The second-largest topic, \textit{sentiment, sentiment analysis, detection, social} (Topic 1), comprises 614 papers (8.6\%), reflecting the massive volume of socially-generated data on platforms like Twitter, and the commercial interest in understanding Arabic public opinion. The remaining topics are significantly smaller, covering specialized domains like medical questionnaires (Topic 2), regional studies (Algerian/Tunisian; Topic 3), specific linguistic tasks such as Named Entity Recognition (NER; Topic 4) and dialect identification (Topic 5), translation of cultural concepts (Topic 6), historical Syriac translation (Topic 7), POS tagging (Topic 8), COVID-19 related research (Topic 9), sarcasm detection (Topic 10), chatbots (Topic 11), gender studies (Topic 12), negation (Topic 13), stress patterns (Topic 14), metaphors (Topic 15), steganography (Topic 16), hallucination in LLMs (Topic 17), and BERT legal applications (Topic 18). The long tail of topics illustrates the breadth of Arabic NLP research, with many niche areas receiving attention from dedicated communities.

\begin{table}[htbp]
\centering
\caption{BERTopic topics with paper counts (after reduction to 20 topics; topic -1 omitted). The outlier topic (-1) contains 2,535 papers (35.6\% of the corpus); the sum of the listed topics is 4,585.}
\label{tab:topics}
\begin{tabular}{clr}
\toprule
\textbf{Topic ID} & \textbf{Topic Name} & \textbf{Count} \\
\midrule
0 & text, speech, translation, recognition & 2,942 \\
1 & sentiment, sentiment analysis, detection, social & 614 \\
2 & patients, version, validity, questionnaire & 305 \\
3 & algerian, algeria, french, dialect & 137 \\
4 & ner, entity, named, named entity & 107 \\
5 & similarity, identification, relatedness, dialect identification & 65 \\
6 & translation, cultural, translators, translating & 63 \\
7 & syriac, greek, translation, he & 60 \\
8 & pos, tagging, tagger, pos tagging & 54 \\
9 & covid, covid 19, 19, pandemic & 52 \\
10 & sarcasm, sarcasm detection, detection, sarcastic & 36 \\
11 & chatbots, chatbot, conversational, medical & 27 \\
12 & women, saudi, gender, female & 27 \\
13 & negation, sentential, sentential negation, verbal & 24 \\
14 & stress, syllable, stress patterns, word & 20 \\
15 & metaphors, metaphor, conceptual, conceptual metaphors & 15 \\
16 & steganography, secret, hiding, text & 14 \\
17 & hallucination, factual, llms, hallucinations & 13 \\
18 & bert, legal, bidirectional, bidirectional encoder & 10 \\
\bottomrule
\end{tabular}
\end{table}

\begin{figure}[htbp]
\centering
\includegraphics[width=0.85\textwidth]{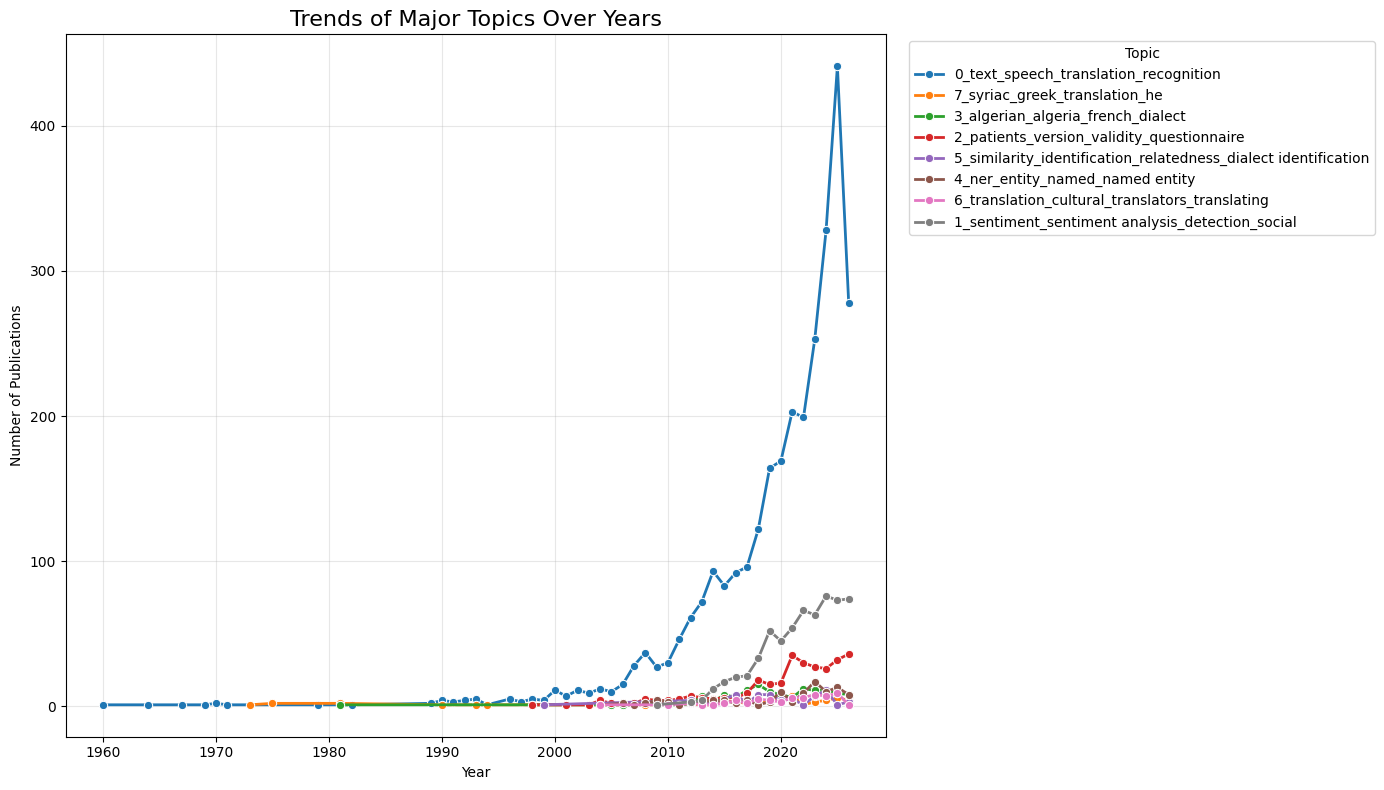}
\caption{Trends of major topics over years.}
\label{fig:topic_trends}
\end{figure}

Figure~\ref{fig:topic_trends} shows the trends of major topics over years. The largest topic (Topic 0) demonstrates a consistent exponential growth trajectory, particularly accelerating after 2020, culminating in a sharp peak in 2025. The abrupt decline observed in 2026 is likely a data collection artifact, as the final year of the dataset is incomplete at the time of analysis. This suggests that as the quality and availability of pre-trained models improves, the barrier to entry for ASR and MT research lowers, leading to a massive output of papers. Sentiment analysis (Topic 1) experienced a sharp rise peaking around 2022--2023 before plateauing and slightly declining. This saturation likely indicates that the basic sentiment classification problem is considered largely "solved" for standard dialects, prompting researchers to move towards more complex tasks or specific sub-domains. The decline in sentiment analysis papers after 2023 may also reflect the shift toward generative LLMs, which can perform sentiment analysis as a zero-shot task, reducing the need for specialized models. The patient-related topic (Topic 2) shows a notable, steady growth after 2019, reflecting the increasing interest in health-related applications and clinical questionnaire validation within Arabic NLP.

\subsubsection{Topic Quality Metrics}

BERTopic significantly outperformed LDA across both quality metrics (Table~\ref{tab:topic_quality}). BERTopic achieved a topic diversity of 0.868 and a coherence of 0.772, compared to 0.720 and 0.028 for LDA. This stark contrast is not surprising, as BERTopic leverages the contextual semantic representations of Transformer models to extract themes, whereas traditional LDA relies on strict bag-of-words assumptions. The exceptionally low coherence score of LDA (0.028) suggests it struggled to identify semantically meaningful word groupings, producing noisy and highly overlapping topics that would be of little use for further scientific analysis. The low diversity of LDA also implies that its topics share many common high-frequency words, reducing their interpretability. Given these results, all subsequent topic-related analyses (citation by topic, trend analysis, task--dialect matrix) are based on the BERTopic output, which provides a more reliable and granular partition of the literature.

\begin{table}[htbp]
\centering
\caption{Topic quality metrics: BERTopic vs. LDA.}
\label{tab:topic_quality}
\begin{tabular}{lcc}
\toprule
\textbf{Metric} & \textbf{BERTopic} & \textbf{LDA} \\
\midrule
Topic Diversity & 0.868 & 0.720 \\
Topic Coherence & 0.772 & 0.028 \\
\bottomrule
\end{tabular}
\end{table}


\subsection{Citation Analysis}

We analyzed citation patterns from multiple angles: most cited papers, H-index per topic, and average citations per topic. These metrics provide complementary views of impact.

\subsubsection{Most Cited Papers}

Table~\ref{tab:top_cited} lists the top-10 most cited papers. The most cited paper is the AraBERT paper \citep{antoun2020arabert}, with 1,480 citations. This is highly expected, as the paper introduces the foundational resource that the majority of subsequent deep-learning research in Arabic is built upon. The rest of the list includes seminal textbooks on Arabic linguistics \citep{holes2004modern, watson2007phonology}, foundational NLP surveys, and crucial tool-building papers such as Arabic tokenization and morphological disambiguation \citep{habash-rambow-2005-arabic}, the MADAMIRA toolkit \citep{pasha-etal-2014-madamira}, and AraVec \citep{soliman2017aravec}. Classical works on Arabic structure and varieties remain heavily cited because they provide the linguistic grounding for computational models, while resources like the Arabic validation of the Patient Health Questionnaire \citep{alhadi2017arabic} and foundational work on translation techniques \citep{molina2004translation} reflect the field's interdisciplinary reach into health and translation studies. The presence of textbooks and linguistic reference works alongside NLP tool papers indicates that the field draws heavily on both linguistic theory and computational resources. The high citation counts of these works also reflect their role as standard references that are cited across many different sub-areas, from speech recognition to sentiment analysis.

\begin{table}[htbp]
\centering
\caption{Top-10 most cited papers.}
\label{tab:top_cited}
\begin{tabular}{clr}
\toprule
\textbf{Rank} & \textbf{Paper} & \textbf{Citations} \\
\midrule
1 & AraBERT: Transformer-based Model for Arabic Language Understanding & 1,480 \\
2 & Translation Techniques Revisited & 1,023 \\
3 & Modern Arabic: Structures, Functions and Varieties (Holes) & 771 \\
4 & ARBERT \& MARBERT: Deep Bidirectional Transformers for Arabic & 714 \\
5 & The Phonology and Morphology of Arabic & 688 \\
6 & Arabic Natural Language Processing (survey) & 551 \\
7 & Arabic Tokenization, POS Tagging and Morphological Disambiguation & 534 \\
8 & Arabic Translation of Patient Health Questionnaire & 523 \\
9 & AraVec: A Set of Arabic Word Embedding Models & 520 \\
10 & The Phonetics and Phonology of Semitic Pharyngeals & 405 \\
\bottomrule
\end{tabular}
\end{table}

The correlation between paper age and citations is positive and statistically significant (r = 0.245, p < 0.001). This confirms that while the field is currently growing rapidly, the foundational literature published over a decade ago continues to accrue citations, serving as the theoretical and practical bedrock for newer research. However, the moderate magnitude of the correlation (0.245) indicates that age is not the sole determinant of citation impact; some recent papers, especially those related to LLMs, have accumulated citations at an extraordinarily fast rate, challenging the traditional age--citation relationship. For instance, papers on Arabic LLMs published in 2023--2024 may already have hundreds of citations, underscoring the accelerated citation dynamics in the current era.

\subsubsection{H-index per Topic}

\begin{figure}[htbp]
\centering
\includegraphics[width=0.8\textwidth]{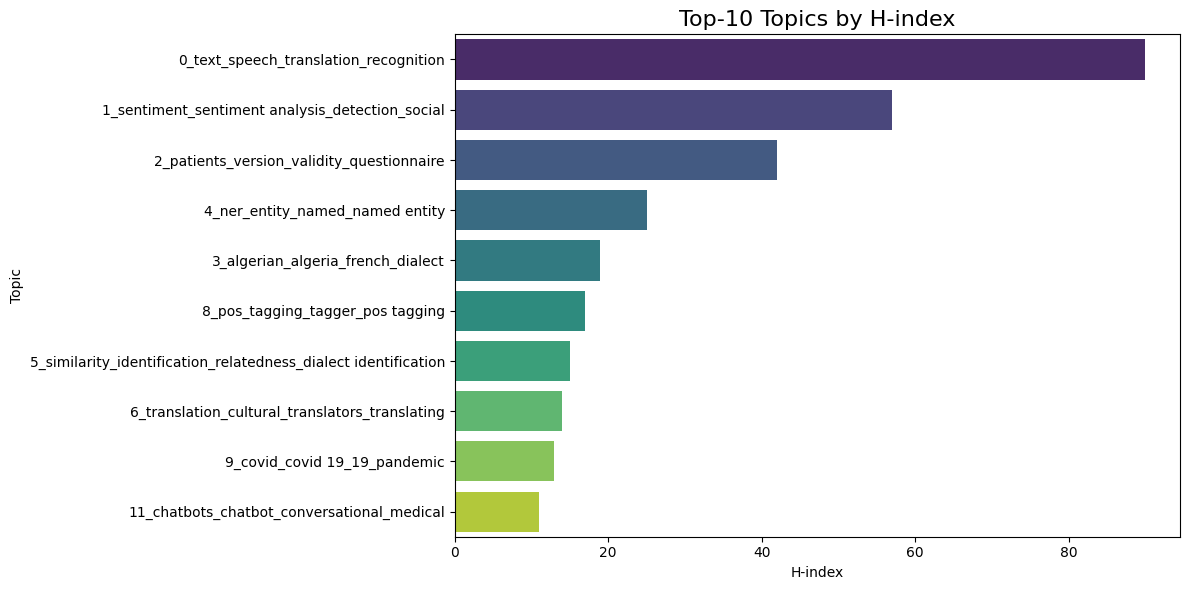}
\caption{Top-10 topics by H-index.}
\label{fig:hindex}
\end{figure}

Figure~\ref{fig:hindex} presents the H-index per topic. The largest topic, \textit{text, speech, translation, recognition}, has the highest H-index (90), followed by sentiment analysis (57) and patient-related studies (42). The H-index for Topic 0 confirms that this volume of research is not just numerous, but also highly impactful; it contains a huge number of papers that have each been cited dozens of times. The elevated H-index for the sentiment analysis topic reflects its historical popularity and widespread commercial application, as many papers on Arabic sentiment analysis have been cited by both academic and industry researchers. The relatively high H-index of the patient-related topic (Topic 2) is noteworthy, as it suggests that medical and health-related Arabic NLP studies, despite being fewer in number, have had a significant impact on clinical applications and cross-cultural validation studies.

\subsubsection{Average Citations by Topic}

\begin{figure}[htbp]
\centering
\includegraphics[width=0.85\textwidth]{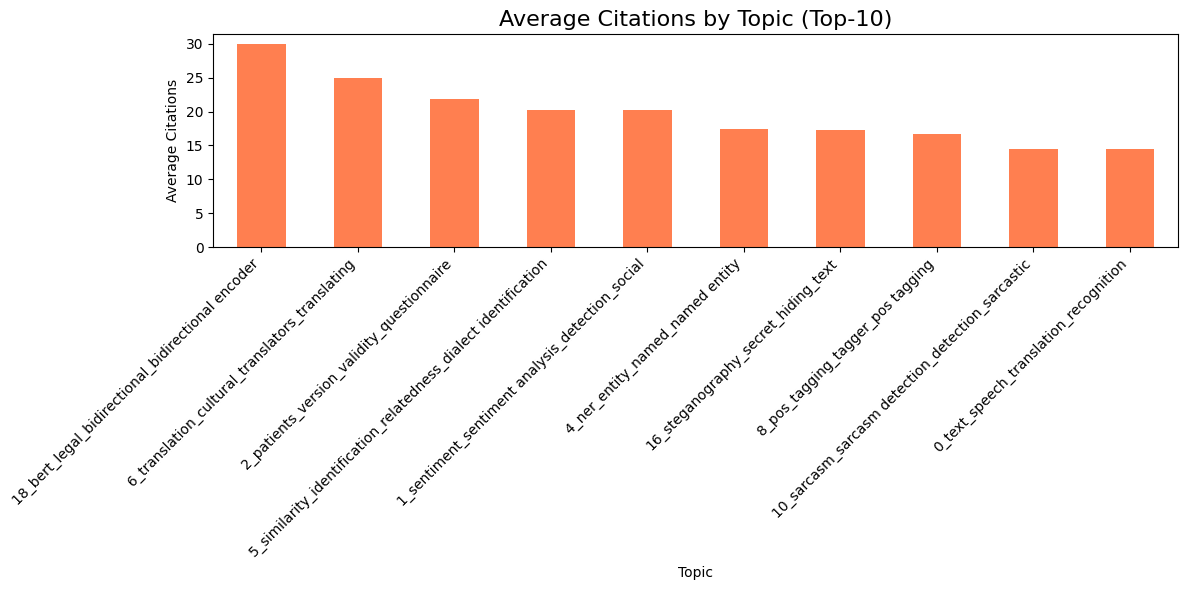}
\caption{Average citations by topic (Top-10).}
\label{fig:avg_citations}
\end{figure}

Figure~\ref{fig:avg_citations} shows the average citations per topic, providing a different perspective on impact. Interestingly, Topic 18 (bert, legal, bidirectional, bidirectional encoder) has the highest average citations (29.90), followed by Topic 6 (translation, cultural, translators, translating) with 24.90 and Topic 2 (patients, version, validity, questionnaire) with 21.84. This is a fascinating finding: while Topic 0 dominates in sheer volume, it does not have the highest average impact. This suggests that highly specialized topics, such as medical questionnaire translation or legal BERT applications, produce fewer papers, but these papers are of such high utility to specific academic communities that they receive higher per-paper citation rates. This highlights a "high-volume, moderate-impact" dynamic for core NLP tasks versus a "low-volume, high-impact" dynamic for specialized domain applications. The high average citations of these niche topics may be due to the scarcity of works in those areas, making each contribution highly cited by researchers seeking foundational references.

\subsection{Regression Analysis: Factors Affecting Citation Impact}

To identify the factors that influence citation counts, we conducted an ordinary least squares (OLS) regression with citation count as the dependent variable. The model included dummy variables for source databases, centered publication year (\texttt{year\_c}), matched term count, institutional presence, and number of authors. The model was statistically significant (F = 92.83, p < 0.001) and explained 10.5\% of the variance in citation counts ($R^2 = 0.105$). Although the explanatory power is modest, in bibliometric studies such values are common due to the high stochasticity of citation behavior. Table~\ref{tab:regression} presents the significant predictors.

\begin{table}[htbp]
\centering
\caption{Significant predictors of citation count (p < 0.05).}
\label{tab:regression}
\begin{tabular}{lrr}
\toprule
\textbf{Predictor} & \textbf{Coefficient} & \textbf{p-value} \\
\midrule
year\_c & -1.234 & <0.001 \\
source\_Semantic\_Scholar & +5.455 & <0.001 \\
source\_OpenAlex & +11.086 & <0.001 \\
has\_institutions & +8.730 & <0.001 \\
source\_OpenAlex\_(extra) & -9.042 & 0.004 \\
matched\_terms\_count & +1.208 & 0.044 \\
\bottomrule
\end{tabular}
\end{table}

Key findings from the regression analysis reveal several critical insights:

\begin{enumerate}
    \item \textbf{Publication year} has a negative coefficient (-1.234), indicating that older papers receive more citations. This is a standard bibliometric phenomenon: older papers have had more time to accumulate citations. The coefficient implies that, on average, a paper published one year earlier receives about 1.23 more citations, holding other factors constant. This effect is highly significant (p < 0.001) and underscores the importance of time in citation accrual.
    \item \textbf{OpenAlex} and \textbf{Semantic Scholar} sources are associated with significantly higher citation counts (+11.09 and +5.45 citations, respectively). This may indicate that these databases have historically indexed higher-quality or more widely circulated venues, or that their indexing strategies preferentially select high-impact works. It could also reflect that papers indexed in these databases are more visible to the research community, leading to higher citation rates.
    \item \textbf{Institutional affiliation} is a strong positive predictor (+8.73 citations). This highlights the importance of institutional prestige and collaborative networks; papers originating from well-funded, active research groups are more visible and, consequently, more cited. The presence of complete institutional metadata may also serve as a proxy for publication quality, as well-maintained metadata often correlates with reputable journals and conferences.
    \item \textbf{Matched terms count} is positively associated with citation count (+1.21 citations per additional matched term). This suggests that papers covering a broader range of topics or specific tasks (operationalized as matches to our term list) are more likely to serve as a bridge for other researchers, thereby increasing their citability. Each additional matched term may signal greater relevance to multiple sub-communities within Arabic NLP, broadening the potential citing audience.
    \item \textbf{is\_OpenAlex\_(extra)} shows a significant negative coefficient (-9.042, p = 0.004). This indicates that papers indexed in the "extra" version of OpenAlex (perhaps a secondary or less curated subset) tend to receive fewer citations compared to papers not in that source. This may be due to the inclusion of lower-quality or less visible works in that source, or it could reflect a metadata artifact. This finding warrants further investigation into the composition of the OpenAlex (extra) subset.
\end{enumerate}

Despite the significance of several predictors, the low R² value suggests that citation counts are influenced by many unobserved factors, such as author reputation, journal prestige, topic popularity at the time of publication, and random chance. Additionally, the extremely skewed distribution of citation counts (as evidenced by the high skewness and kurtosis in the full model output) violates the normality assumption of OLS. Therefore, we recommend that future work employ count-based regression models (e.g., negative binomial or Poisson) or use log-transformed citation counts to better handle the overdispersion. Nonetheless, the current analysis provides a useful first approximation of the factors associated with citation impact in Arabic NLP.

\subsection{Geographic Distribution}

\begin{figure}[htbp]
\centering
\includegraphics[width=0.9\textwidth]{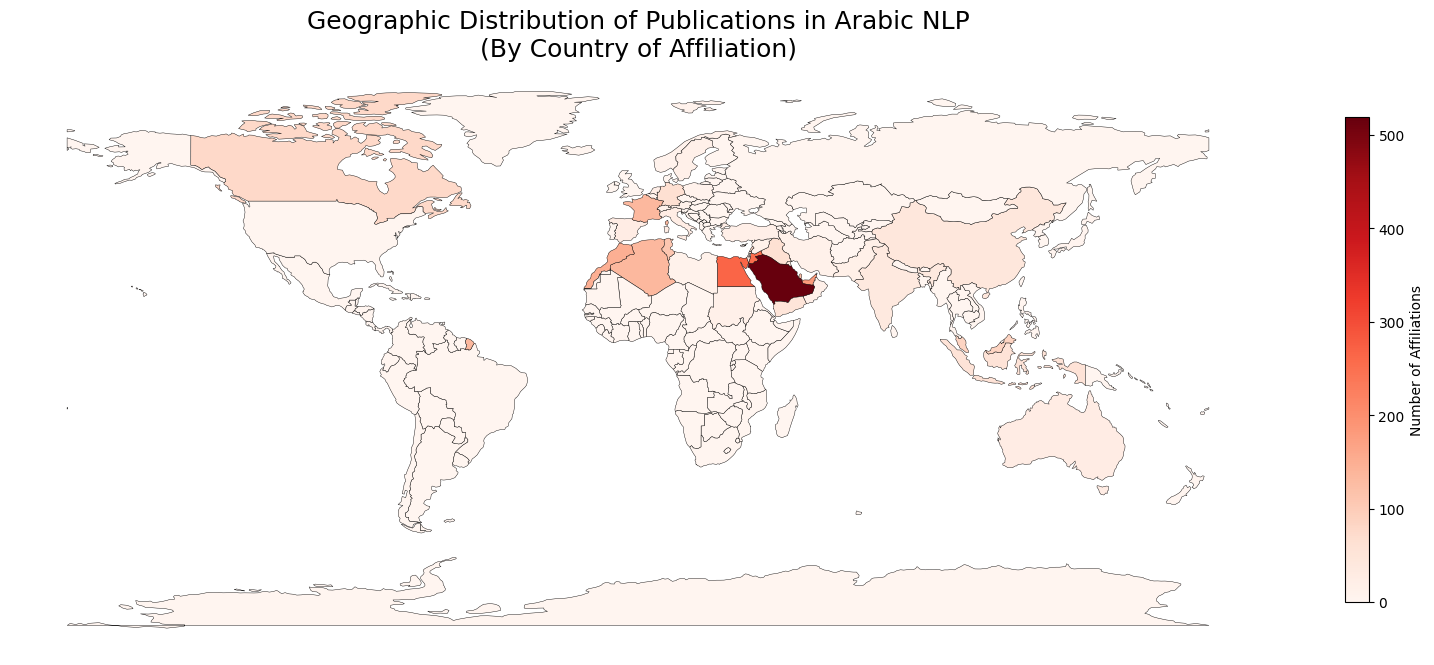}
\caption{Geographic distribution of Arabic NLP publications by country of affiliation.}
\label{fig:worldmap}
\end{figure}

Figure~\ref{fig:worldmap} shows the geographic distribution of Arabic NLP publications by country of affiliation. Saudi Arabia (519 affiliations), the United States (463), and Egypt (266) are the top three contributing countries. Saudi Arabia's top position is largely fueled by massive state-backed funding initiatives for AI and Natural Language Processing, most notably through King Abdulaziz City for Science and Technology (KACST) and major universities like King Saud University and KAUST. The United States, meanwhile, represents a hub of globally-minded researchers and prestigious universities (e.g., Columbia, CMU) that frequently collaborate with Arabic-speaking researchers. Egypt's strong showing reflects its large academic community and historical leadership in Arabic linguistics and computer science.

Table~\ref{tab:top_countries} lists the top-10 countries by affiliation count. The strong presence of countries like Egypt, Jordan, the UK, and the UAE underscores that Arabic NLP is a truly global endeavor, with a network of researchers working both within Arabic-speaking nations and in diaspora communities across Europe and North America. The relatively high numbers for Morocco, France, and Algeria also point to a distinct Francophone research cluster focusing on Maghrebi dialects, reflecting historical and linguistic ties. The United Kingdom's presence is notable, driven by institutions like the University of Edinburgh and the University of Sheffield, which have strong NLP groups with an interest in Arabic. The United Arab Emirates and Qatar have invested heavily in AI research, as evidenced by their top-10 positions and the presence of institutions like the Mohamed bin Zayed University of Artificial Intelligence and Qatar University.

\begin{table}[htbp]
\centering
\caption{Top-10 countries by affiliation count.}
\label{tab:top_countries}
\begin{tabular}{lrr}
\toprule
\textbf{Country} & \textbf{Code} & \textbf{Affiliations} \\
\midrule
Saudi Arabia & SA & 519 \\
United States & US & 463 \\
Egypt & EG & 266 \\
Jordan & JO & 253 \\
United Kingdom & GB & 212 \\
United Arab Emirates & AE & 182 \\
Morocco & MA & 150 \\
France & FR & 134 \\
Algeria & DZ & 134 \\
Qatar & QA & 123 \\
\bottomrule
\end{tabular}
\end{table}

It is important to note that the country affiliation counts are based on the subset of papers with available metadata (36.1\% of the corpus). Therefore, these rankings may be biased toward countries with more complete metadata standards, typically wealthier institutions and those in Western or Gulf countries. The actual global distribution may include a larger share from other Arabic-speaking countries (e.g., Iraq, Sudan, Yemen) that are underrepresented due to metadata gaps. Nevertheless, the observed distribution highlights the concentration of research capacity in a few resource-rich nations, leaving many Arabic-speaking regions underserved in terms of local NLP research infrastructure.


\subsection{Institutional Distribution}

\begin{figure}[htbp]
\centering
\includegraphics[width=0.85\textwidth]{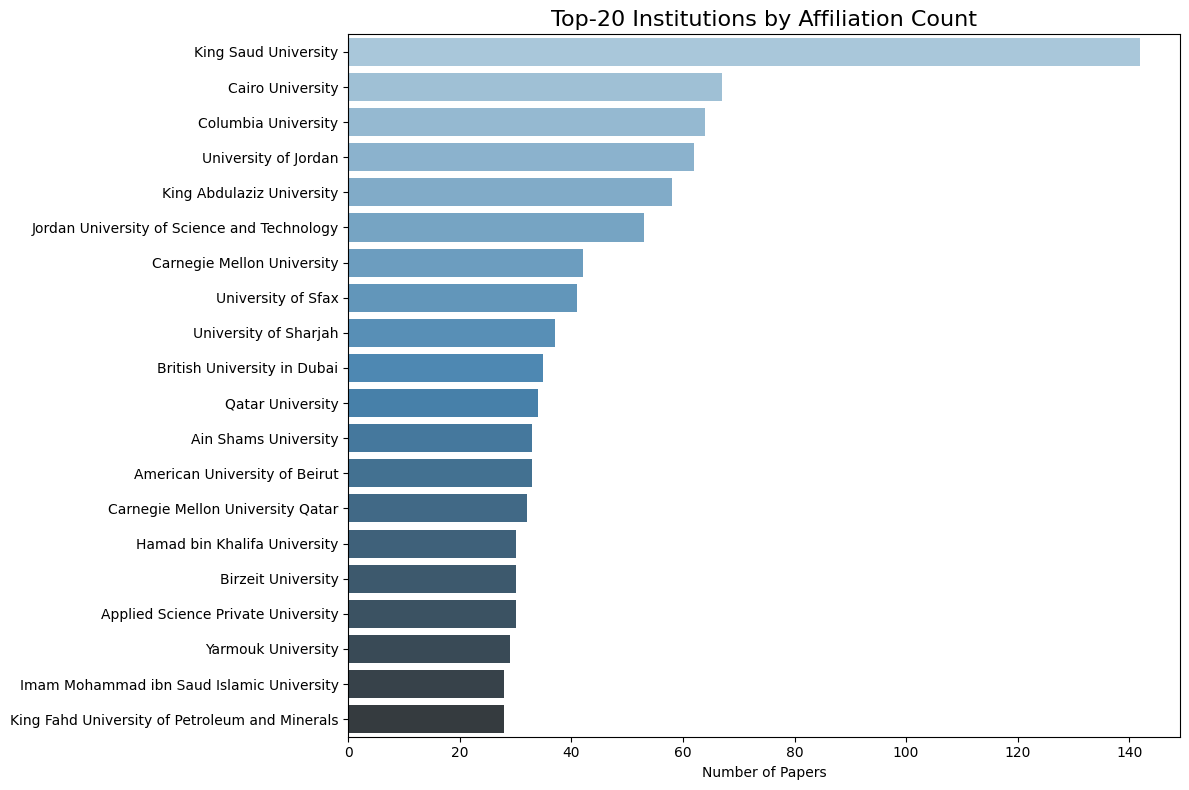}
\caption{Top-20 institutions by affiliation count.}
\label{fig:institutions}
\end{figure}

Figure~\ref{fig:institutions} shows the top-20 institutions by affiliation count. King Saud University (142 papers) is the most productive institution, followed by Cairo University (67) and Columbia University (64). This distribution closely mirrors the country-level analysis and highlights the immense output contributed by highly centralized, well-funded institutions in the Middle East. King Saud University's dominance is particularly striking, as it accounts for nearly 6\% of all papers with affiliation data, driven by its active NLP research group and strong governmental support for AI research.

The rest of the list is dominated by a mix of major global universities (e.g., Carnegie Mellon, Qatar University, Columbia) and regional powerhouses (e.g., University of Jordan, King Abdulaziz University). This distribution indicates that Arabic NLP research is highly concentrated in a select group of elite institutions that possess the computational infrastructure, funding, and talent pools necessary for advanced deep learning research. The presence of several institutions in the UAE and Qatar also points to a highly competitive and well-resourced research environment in the Gulf region. Notably, Carnegie Mellon University appears both in its main campus (42 papers) and its Qatar branch (32 papers), reflecting the strategic expansion of top-tier NLP programs into the Gulf. The inclusion of universities from non-Arab countries (e.g., University of Sfax, Tunisia; Applied Science Private University, Jordan) further demonstrates the regional diversity of productive institutions.

A closer look at the top institutions reveals different research profiles. For example, Cairo University and Ain Shams University have strong traditions in Arabic linguistics and rule-based NLP, while Columbia University and Carnegie Mellon are known for their contributions to statistical and neural methods, often in collaboration with Arabic-speaking researchers. This institutional diversity enriches the field by combining deep linguistic expertise with cutting-edge computational techniques.


\subsection{Dialect Analysis}

We analyzed the presence of different Arabic dialects in the abstracts of the papers to understand the geographical and linguistic focus of research. This analysis is crucial because Arabic is a diglossic language with a significant gap between Modern Standard Arabic (MSA) and the various regional dialects.

\subsubsection{Dialect Mentions Over Years}

\begin{figure}[htbp]
\centering
\includegraphics[width=0.85\textwidth]{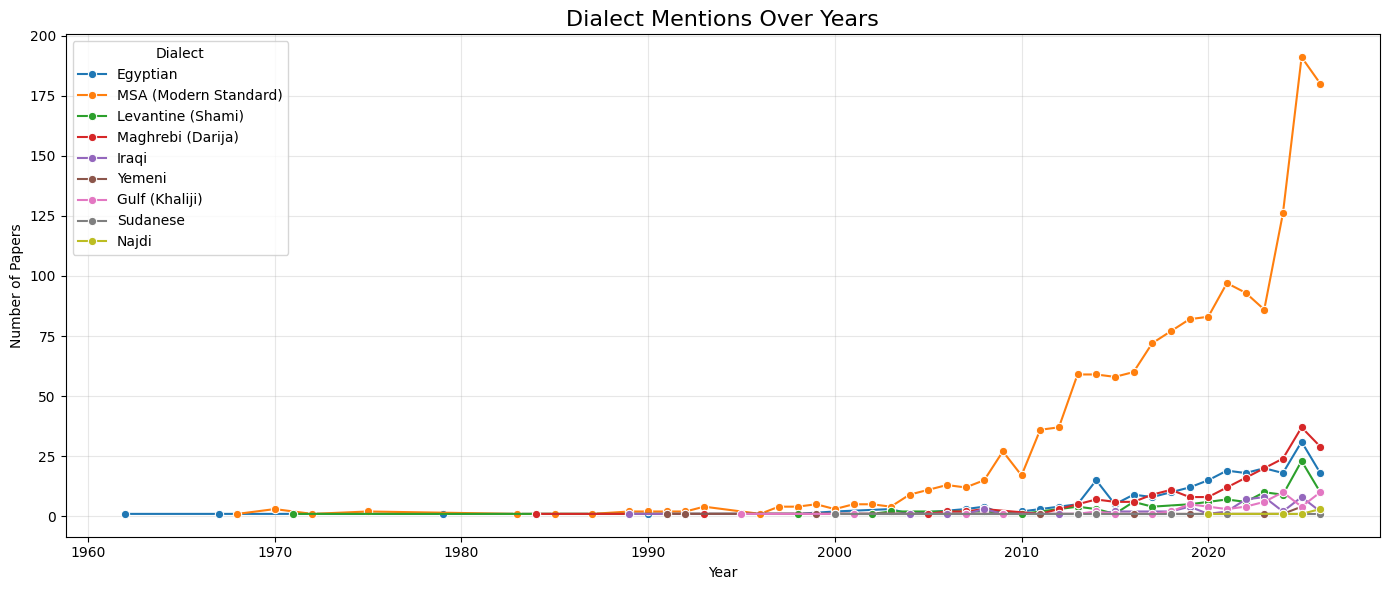}
\caption{Dialect mentions over years.}
\label{fig:dialect_trends}
\end{figure}

Figure~\ref{fig:dialect_trends} shows the evolution of dialect mentions over time. Modern Standard Arabic (MSA) consistently dominates, with a sharp increase after 2020. MSA's dominance is expected, as it is the official written standard used in news, education, and official documents, providing a stable target for NLP systems. The post-2020 spike is particularly pronounced, reaching a peak of approximately 190 mentions in 2025 before a slight decline in 2026, which is likely attributable to incomplete data for the final year. This surge is driven by the increased availability of large MSA corpora and the focus of LLM development on high-resource languages, where MSA is considered the primary variant.

Egyptian Arabic shows fluctuating growth with distinct peaks, notably around 2015 and again after 2020, peaking at roughly 25 mentions in 2025. Its popularity stems from its massive media industry and historical influence, making it the most resource-rich dialect for research. In contrast, Maghrebi (Darija) exhibits a more consistent upward trajectory, accelerating after 2020 and reaching approximately 35 mentions by 2024. This growth indicates a rising awareness of the vast under-resourced status of North African dialects, driven largely by research communities in France and Morocco.

Table~\ref{tab:dialect_mentions} presents the total counts for each dialect. MSA is the most frequently mentioned (1,553 mentions), followed by Egyptian (462) and Maghrebi/Darija (341). The low counts for dialects like Hejazi (1), Sudanese (20), and Yemeni (28) highlight a massive gap in resource allocation. These dialects suffer from a severe lack of corpora and linguistic tools, leaving them largely marginalized in the current NLP ecosystem. The near-total absence of Hassaniya (0 mentions) is particularly concerning, as it indicates that the dialect spoken in Mauritania and parts of Western Sahara is entirely unrepresented in the literature, making it a critical area for future research. It is important to note that these counts reflect mentions in abstracts, which may undercount work that focuses on a dialect without naming it explicitly (e.g., using terms like "Maghrebi Arabic" or "North African dialect"). Nevertheless, the relative proportions clearly show the dominance of MSA and the under-representation of most dialects.

\begin{table}[htbp]
\centering
\caption{Dialect mentions in abstracts (total counts).}
\label{tab:dialect_mentions}
\begin{tabular}{lr}
\toprule
\textbf{Dialect} & \textbf{Mentions} \\
\midrule
MSA (Modern Standard) & 1,553 \\
Egyptian & 462 \\
Maghrebi (Darija) & 341 \\
Levantine (Shami) & 187 \\
Gulf (Khaliji) & 105 \\
Iraqi & 79 \\
Najdi & 35 \\
Yemeni & 28 \\
Sudanese & 20 \\
Hejazi & 1 \\
Hassaniya & 0 \\
\bottomrule
\end{tabular}
\end{table}

\subsubsection{Dialects and Major Topics: Trends Over Years}

\begin{figure}[htbp]
\centering
\includegraphics[width=0.85\textwidth]{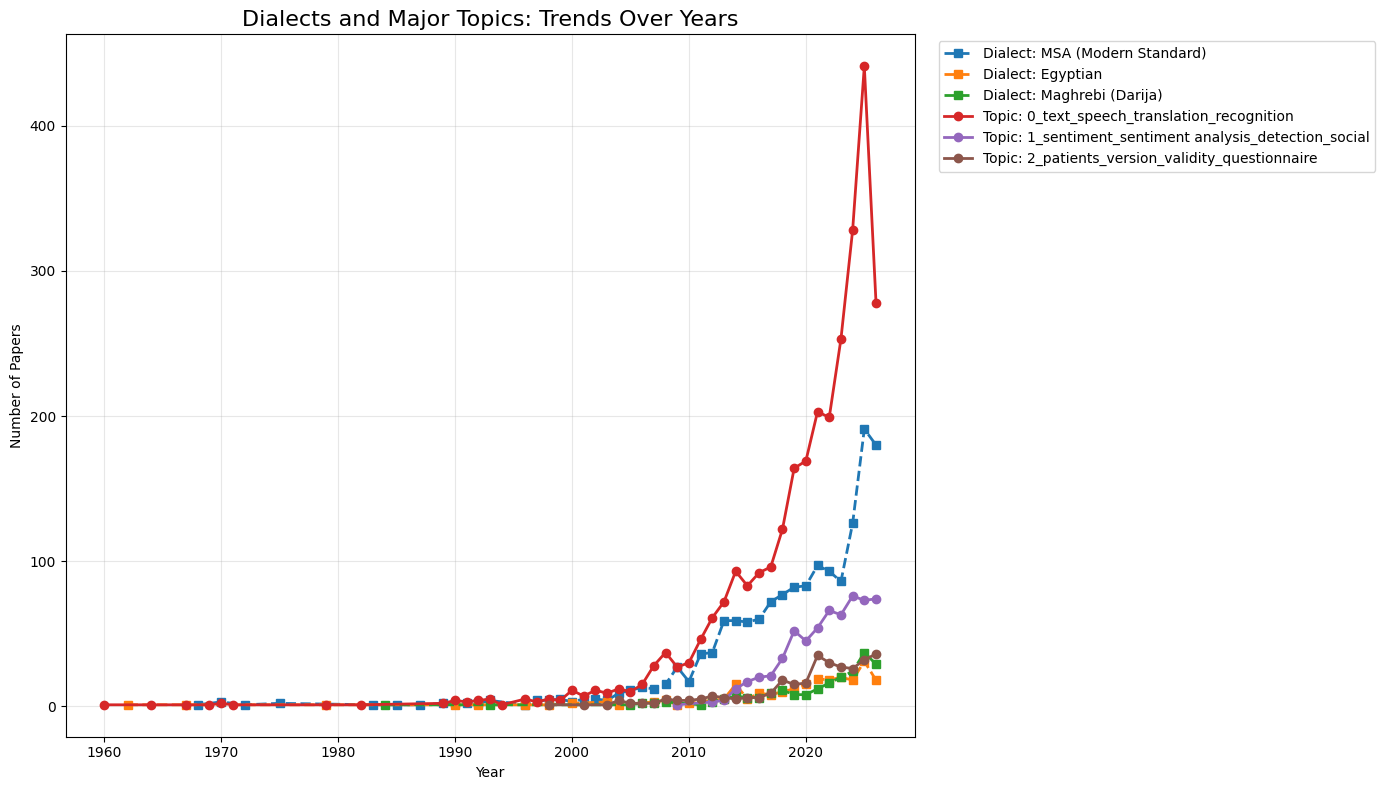}
\caption{Dialects and major topics: trends over years.}
\label{fig:dialect_topic_overlay}
\end{figure}

Figure~\ref{fig:dialect_topic_overlay} overlays dialect mentions with major topic trends. MSA correlates strongly with the largest topic (Topic 0: text, speech, translation, recognition), following a broadly similar upward trajectory, although Topic 0 reaches a much higher peak (around 450 papers in 2025) compared to MSA (around 190 papers in the same year). This confirms that most core NLP research (translation, ASR, OCR) is still fundamentally focused on MSA. Egyptian and Maghrebi dialects show alignment with sentiment analysis and patient-related studies, exhibiting a moderate rise peaking around 2023--2024. This suggests that social media sentiment analysis and health-related applications are the primary domains driving research into these specific dialects, as they directly address the language needs of their respective populations. For instance, Egyptian Arabic is widely used in social media, making it a natural testbed for sentiment analysis, while Maghrebi dialects are often studied in the context of code-switching and patient communication in France.

The task--dialect gap matrix (Table~\ref{tab:dialect_gaps} in Section 4.11) reveals that certain tasks, such as NER and POS tagging, are almost exclusively associated with MSA, with little to no dialect-specific research. This gap is particularly problematic because dialectal text often exhibits greater morphological and orthographic variability, making existing MSA-trained models perform poorly. The lack of research on dialectal NER, for example, hinders the development of downstream applications like information extraction from dialectal social media posts.


\subsection{Task--Dialect Gap Matrix}

To systematically identify research gaps at the intersection of specific NLP tasks and Arabic dialects, we constructed a task--dialect matrix. The matrix cross-tabulates the percentage of papers (row-normalized) that mention both a given task and a given dialect in their abstracts. Table~\ref{tab:dialect_gaps} presents the results for the five most commonly studied tasks, while the remaining tasks are discussed separately due to their distinct characteristics. This analysis is crucial for pinpointing areas where no or very little research has been conducted, thereby guiding future resource allocation and research priorities.

\begin{table}[htbp]
\centering
\footnotesize
\setlength{\tabcolsep}{3.5pt}
\caption{Task--dialect coverage matrix for core NLP tasks (row-normalized, \%).}
\label{tab:dialect_gaps}
\begin{tabular}{lrrrrrrrrrrr}
\toprule
\textbf{Task} & \textbf{MSA} & \textbf{Egyptian} & \textbf{Levantine} & \textbf{Gulf} & \textbf{Maghrebi} & \textbf{Iraqi} & \textbf{Sudanese} & \textbf{Yemeni} & \textbf{Najdi} & \textbf{Hejazi} & \textbf{Hassaniya} \\
\midrule
sentiment & 53.9 & 16.6 & 2.4 & 4.7 & 15.9 & 2.7 & 2.7 & 0.3 & 0.7 & \textbf{0.0} & \textbf{0.0} \\
translation & 55.3 & 19.3 & 5.5 & 4.6 & 10.7 & 3.8 & 0.2 & \textbf{0.0} & 0.6 & \textbf{0.0} & \textbf{0.0} \\
ner & 58.0 & 15.7 & 6.1 & 4.0 & 10.4 & 2.6 & 0.4 & 1.2 & 1.5 & \textbf{0.0} & \textbf{0.0} \\
asr & 51.3 & 24.2 & 5.1 & 4.0 & 11.2 & 2.5 & 0.4 & 0.4 & 1.1 & \textbf{0.0} & \textbf{0.0} \\
dialect\_id & 56.0 & 26.6 & 4.3 & 2.7 & 8.7 & 0.5 & \textbf{0.0} & \textbf{0.0} & 1.1 & \textbf{0.0} & \textbf{0.0} \\
\bottomrule
\end{tabular}
\end{table}

The matrix reveals extreme resource disparity: MSA is present in over 50\% of papers across all tasks, indicating that these tasks are almost exclusively studied using MSA data. Egyptian Arabic is the second most studied dialect, with double-digit percentages across most tasks, thanks to its high media presence and linguistic resources. In stark contrast, dialects such as Hejazi and Hassaniya show \textbf{zero coverage} across all five tasks, while Sudanese and Yemeni appear only sporadically. These zero-coverage pairs highlight strategic opportunities for future research, particularly for these under-resourced dialects. There is a critical and urgent need for the community to move beyond the "low-hanging fruit" of MSA and Egyptian and invest in collecting corpora and developing foundational language models for these marginalised dialects. Prioritising these gaps is essential not only for academic completeness but for ensuring equitable access to modern language technologies for all Arabic speakers. For example, the complete absence of sentiment analysis for Hejazi and Hassaniya means that sentiment tools cannot be deployed for social media monitoring in western Saudi Arabia or Mauritania, limiting real-world applications in those regions.

\subsection*{Specialised Tasks: OCR and Summarisation}

Beyond the core tasks presented in Table~\ref{tab:dialect_gaps}, two tasks exhibit distinctive patterns that warrant separate discussion: OCR and summarisation.

\begin{table}[htbp]
\centering
\footnotesize
\setlength{\tabcolsep}{3.5pt}
\caption{Task--dialect coverage for specialised tasks (row-normalized, \%).}
\label{tab:specialised_tasks}
\begin{tabular}{lrrrrrrrrrrr}
\toprule
\textbf{Task} & \textbf{MSA} & \textbf{Egyptian} & \textbf{Levantine} & \textbf{Gulf} & \textbf{Maghrebi} & \textbf{Iraqi} & \textbf{Sudanese} & \textbf{Yemeni} & \textbf{Najdi} & \textbf{Hejazi} & \textbf{Hassaniya} \\
\midrule
ocr & 37.5 & 25.0 & 4.2 & \textbf{0.0} & \textbf{33.3} & \textbf{0.0} & \textbf{0.0} & \textbf{0.0} & \textbf{0.0} & \textbf{0.0} & \textbf{0.0} \\
summarisation & 76.9 & 15.4 & \textbf{0.0} & 7.7 & \textbf{0.0} & \textbf{0.0} & \textbf{0.0} & \textbf{0.0} & \textbf{0.0} & \textbf{0.0} & \textbf{0.0} \\
\bottomrule
\end{tabular}
\end{table}

As shown in Table~\ref{tab:specialised_tasks}, OCR research has a surprisingly high share of Maghrebi (33.3\%), likely due to specific projects on digitising Maghrebi manuscripts, but \textbf{zero coverage} for Gulf, Iraqi, Sudanese, Yemeni, Najdi, Hejazi, and Hassaniya. This suggests that OCR efforts are highly localised and driven by specific historical or cultural preservation projects rather than systematic coverage. Similarly, summarisation is heavily skewed towards MSA (76.9\%), with Egyptian (15.4\%) and a small Gulf presence (7.7\%), reflecting the general scarcity of dialectal summarisation datasets. All other dialects (Levantine, Maghrebi, Iraqi, Sudanese, Yemeni, Najdi, Hejazi, and Hassaniya) show zero coverage for summarisation.

These gaps underscore the need for a more balanced research agenda that considers the full diversity of Arabic dialects, especially as large language models become increasingly integrated into everyday applications across the Arab world. The stark contrast between the well-resourced MSA and Egyptian dialects and the near-total absence of data for dialects like Hejazi, Hassaniya, and Sudanese represents not just an academic oversight but a fundamental barrier to equitable language technology deployment.


\subsection{Supplementary Statistical Analyses}

In addition to the main analyses, we conducted several supplementary statistical tests to further characterize the temporal and structural evolution of the field.

The chi-squared test for independence between decade and topic distribution was significant ($\chi^2 = 75.04$, df = 8, p < 0.001), indicating that the distribution of research topics differs significantly across decades (1960--2009, 2010--2019, 2020--2026). This confirms the empirical observation that the field has undergone a profound paradigm shift. In the early decades, foundational work on linguistic theory and rule-based processing dominated. In the 2010s, machine learning and deep learning began to take root, and in the 2020s, the focus has decisively shifted to pre-trained Transformers and generative LLMs. The significant chi-squared result suggests that these shifts are not random but reflect systematic changes in research priorities driven by technological advancements.

Table~\ref{tab:cagr} presents the Compound Annual Growth Rates (CAGR) for the top five topics over the last five years. The largest topic (text, speech, translation, recognition) shows the highest growth rate (CAGR = 11.54\%), indicating that core NLP tasks are still expanding at a rapid pace. The patient-related topic (CAGR = 5.63\%) is also growing, reflecting continued interest in health-related applications. Sentiment analysis (CAGR = 3.69\%) shows more moderate growth. In contrast, NER (CAGR = -5.26\%) and the Algerian/Tunisian dialect topic (CAGR = -9.80\%) exhibit negative growth, potentially indicating saturation or a shift in research focus.

\begin{table}[htbp]
\centering
\caption{Compound Annual Growth Rate (CAGR) for top topics (last 5 years).}
\label{tab:cagr}
\begin{tabular}{lr}
\toprule
\textbf{Topic} & \textbf{CAGR (\%)} \\
\midrule
text, speech, translation, recognition & 11.54 \\
patients, version, validity, questionnaire & 5.63 \\
sentiment, sentiment analysis, detection, social & 3.69 \\
ner, entity, named, named entity & -5.26 \\
algerian, algeria, french, dialect & -9.80 \\
\bottomrule
\end{tabular}
\end{table}

These growth rates, while informative, should be interpreted with caution because they are based on a relatively short five-year window and are sensitive to the topic assignment threshold. Nevertheless, they provide a useful snapshot of which research areas are expanding and which are contracting in the current landscape.


\subsection{Concept Analysis}

We also examined the high-level concepts associated with the papers, as provided by the underlying databases (primarily OpenAlex). These concepts are drawn from a controlled vocabulary and reflect the interdisciplinary nature of the field. Table~\ref{tab:concepts} lists the top-20 concepts by frequency.

\begin{table}[htbp]
\centering
\caption{Top-20 concepts by frequency.}
\label{tab:concepts}
\begin{tabular}{lr}
\toprule
\textbf{Concept} & \textbf{Frequency} \\
\midrule
Linguistics & 2,998 \\
Philosophy & 2,899 \\
Arabic & 2,823 \\
Computer science & 2,728 \\
Artificial intelligence & 2,262 \\
Natural language processing & 2,241 \\
Speech recognition & 655 \\
Modern Standard Arabic & 625 \\
Psychology & 618 \\
Mathematics & 528 \\
History & 502 \\
Biology & 502 \\
Programming language & 490 \\
Physics & 395 \\
Sentiment analysis & 379 \\
Task (project management) & 373 \\
World Wide Web & 368 \\
Engineering & 286 \\
Economics & 286 \\
Sociology & 285 \\
\bottomrule
\end{tabular}
\end{table}

The dominance of linguistics and philosophy (often associated with logic and semantics) reflects the strong theoretical grounding of Arabic NLP, where morphological and syntactic analysis play a central role. The high frequency of "Arabic" as a concept is unsurprising. The presence of "Artificial intelligence" and "Natural language processing" at the top confirms the field's alignment with modern AI research. Interestingly, "Speech recognition" appears as a distinct concept, highlighting the importance of spoken language processing in Arabic NLP. The inclusion of "Psychology" and "Biology" likely stems from applications in mental health and biomedical text processing, such as the adaptation of clinical questionnaires and analysis of medical literature. The presence of "History" and "Physics" may be attributed to studies involving historical Arabic manuscripts or acoustic phonetics. Overall, the concept analysis provides a macro-level view of the field's interdisciplinary connections, showing that Arabic NLP draws from, and contributes to, a wide range of scientific domains.


\subsection{Matched Terms Analysis}

As part of our filtering process, we matched papers against a predefined list of Arabic-specific terms. This section analyzes the distribution and impact of these matched terms. Table~\ref{tab:matched_terms} shows the top-20 matched terms by frequency across the corpus.

\begin{table}[htbp]
\centering
\caption{Top-20 matched terms by frequency.}
\label{tab:matched_terms}
\begin{tabular}{lr}
\toprule
\textbf{Matched Term} & \textbf{Frequency} \\
\midrule
arabic & 6,981 \\
arabic language & 2,996 \\
arabic dialect & 1,455 \\
modern standard arabic & 1,329 \\
dialectal arabic & 457 \\
arabic dataset & 447 \\
arabic translation & 426 \\
arabic sentiment & 416 \\
arabic speech & 363 \\
arabic nlp & 316 \\
arabic natural language & 310 \\
arabic corpus & 270 \\
arabic morphology & 145 \\
Standard Arabic & 121 \\
Modern Standard Arabic & 112 \\
arabic llm & 102 \\
arabic dialects & 99 \\
arabic ocr & 87 \\
arabic asr & 85 \\
arabic bert & 74 \\
\bottomrule
\end{tabular}
\end{table}

The distribution of matched terms reflects the research priorities discussed earlier. The high frequency of generic terms like "arabic" and "arabic language" indicates that many papers are broadly focused on the language itself, while more specific terms like "arabic dialect", "modern standard arabic", and "dialectal arabic" reveal the attention to diglossia. The presence of task-specific terms such as "arabic translation", "arabic sentiment", "arabic speech", and "arabic nlp" aligns with the dominant topics identified by BERTopic. The emergence of "arabic llm" (102 papers) and "arabic bert" (74 papers) in the top-20 highlights the recent surge in research on language models for Arabic.

We also computed the correlation between the number of matched terms per paper and its citation count. The correlation was positive but weak (r = 0.036). This suggests that papers covering a broader range of Arabic-specific aspects (i.e., having more matched terms) tend to receive slightly more citations, but the effect is negligible. This finding is consistent with the regression result, where matched terms count had a small but significant positive coefficient (+1.208), indicating that while the effect exists, it is not a strong determinant of citation impact. The weak correlation also implies that highly specialized papers with few matched terms can still achieve high citation counts if they make significant contributions to a specific niche.

Finally, we examined the top matched terms per BERTopic to validate the interpretability of the topics. For instance, Topic 0 (text, speech, translation, recognition) is characterized by terms like "arabic" (2,882), "arabic language" (1,304), and "modern standard arabic" (620); Topic 1 (sentiment) by "arabic sentiment" (198) and "arabic language" (264); and Topic 4 (NER) by "arabic ner" (54). These associations confirm that the topics are semantically coherent and accurately reflect the underlying research themes.


\subsection{Co-authorship Network Analysis}

To understand the collaborative structure of the field, we constructed an undirected co-authorship network and computed degree and betweenness centrality measures. Table~\ref{tab:centrality} lists the seven highest-ranked authors for each measure.

\begin{table}[htbp]
\centering
\caption{Top-7 authors by degree and betweenness centrality.}
\label{tab:centrality}
\begin{tabular}{lrlr}
\toprule
\multicolumn{2}{c}{\textbf{Degree Centrality}} & \multicolumn{2}{c}{\textbf{Betweenness Centrality}} \\
\textbf{Author} & \textbf{Degree} & \textbf{Author} & \textbf{Betweenness} \\
\midrule
Mustafa Jarrar & 0.166 & Nizar Habash & 0.025 \\
Nizar Habash & 0.165 & Marwan Torki & 0.020 \\
Muhammad Abdul-Mageed & 0.163 & Allan Ramsay & 0.019 \\
Omer Nacar & 0.160 & Khaled Shaalan & 0.019 \\
AbdelRahim Elmadany & 0.159 & Bashar Talafha & 0.018 \\
Ismail Berrada & 0.159 & Eiman Alsharhan & 0.018 \\
Preslav Nakov & 0.159 & Ahmed Abdelali & 0.017 \\
\bottomrule
\end{tabular}
\end{table}

Degree centrality reflects the number of direct co-authors and thus the size of an author's immediate collaboration network. The values in Table~\ref{tab:centrality} indicate that the most connected authors have a normalized degree of approximately 0.16, suggesting extensive collaboration activity. Mustafa Jarrar, for example, has the highest degree centrality despite not being the most productive author, which points to his role in connecting different research groups. Nizar Habash and Muhammad Abdul-Mageed also appear among the most central, consistent with their leadership in the field and their involvement in large-scale collaborative projects.

Betweenness centrality captures the extent to which an author serves as a bridge between otherwise disconnected parts of the network. Nizar Habash has the highest betweenness score (0.025), reflecting his role as an intermediary across sub-communities. The presence of researchers such as Marwan Torki, Allan Ramsay, and Khaled Shaalan among the top authors by betweenness centrality further suggests that cross-institutional and cross-disciplinary collaboration is facilitated by a relatively small number of highly connected individuals. The rankings are based on the available metadata and should be regarded as indicative. Overall, the co-authorship network reveals a field in which a core set of well-connected researchers acts as hubs and bridges, facilitating the diffusion of ideas and resources across the community.


\subsection{Summary of Results}

In summary, our analysis reveals a field in rapid transformation. Arabic NLP has grown from a niche area of computational linguistics into a vibrant, multidisciplinary domain fueled by deep learning and, most recently, large language models. The publication output has surged dramatically since 2020, with core tasks like speech recognition, machine translation, and OCR dominating the volume, while sentiment analysis remains a major application area. The field is highly collaborative, with a core set of productive authors and institutions driving progress, but it also exhibits significant disparities: research is heavily concentrated on MSA and Egyptian Arabic, leaving many dialects severely under-resourced. Citation patterns indicate that foundational resources (e.g., AraBERT, Arabic Treebank) have had the greatest impact, but specialized domains such as health and social media also yield high per-paper citations. Regression analysis shows that institutional affiliation, source database, and publication age are significant predictors of citation counts, though much variance remains unexplained. Our topic modeling demonstrates the superiority of BERTopic over LDA for capturing coherent themes, and our task--dialect gap matrix identifies concrete opportunities for future research, particularly in dialectal summarization, OCR, and dialect identification. These findings collectively paint a picture of a dynamic but unevenly developed field, with clear pathways for future growth and equity.


\section{Discussion}\label{sec:discussion}

This section interprets the principal findings, situates them within the broader research landscape, and outlines their implications for different stakeholders. Rather than repeating the numerical results already presented, we focus here on their meaning, their relation to prior work, and the opportunities they reveal.

\subsection{Interpretation of Key Findings}

The exponential growth of Arabic NLP publications---approximately 82\% of all papers in our corpus appeared after 2020---reflects the field's decisive shift toward transformer-based architectures and large language models. This acceleration is not merely a local echo of global NLP trends; it is particularly consequential for Arabic, whose rich morphology, diglossia, and dialectal diversity have historically posed serious obstacles to computational processing \citep{guellil2021arabic, alayba2025arabic}. The fact that such rapid expansion occurred despite these long-standing barriers suggests that modern deep learning methods are finally overcoming them.

A central pattern in our data is the dominance of infrastructure-building research. Topic 0 (\textit{text, speech, translation, recognition}) accounts for 41.3\% of the corpus, underscoring a field still occupied with constructing the tools, datasets, and models that researchers in English NLP often take for granted. This aligns with \citet{iwidat2026arabic}, who noted that resource building represents roughly 21\% of dialectal Arabic NLP work. The high citation counts of foundational resources such as AraBERT, ARBERT \& MARBERT further confirm that creating reusable infrastructure remains one of the most impactful activities in the field. We expect this emphasis to persist until a critical mass of shared resources allows the community to redirect attention toward downstream applications.

Sentiment analysis, the second-largest topic (614 papers, 8.6\%), followed a trajectory of rapid growth beginning around 2017, peaking between 2020 and 2022, and then gradually declining. This pattern mirrors the evolution of sentiment analysis in English NLP, albeit with a delay of two to three years \citep{hamed2025survey}. The early surge was driven by the abundance of Arabic social media data and strong commercial interest in opinion mining \citep{shi2025comprehensive, abo2019sentiment}. The subsequent slowdown likely reflects saturation of basic polarity classification and a shift toward more complex tasks such as aspect-based analysis, irony detection, and emotion recognition \citep{amzil2025sentiment}. This maturation suggests that Arabic sentiment analysis is entering a second wave focused on fine-grained and context-aware tasks.

Perhaps the most consequential finding concerns the severe under-representation of Arabic dialects. Modern Standard Arabic appears in 1,553 papers, while Hejazi is mentioned once, Hassaniya not at all, and Sudanese only 20 times. This disparity stems from the historical emphasis on MSA as the official written standard, but it is increasingly untenable given the prevalence of dialects in everyday communication \citep{dahou2025survey, sakhi2026processing}. The task--dialect gap matrix (Table~\ref{tab:dialect_gaps}) shows that the neglect of dialects is not uniform across tasks: OCR research, for instance, includes a relatively high proportion of Maghrebi work (33.3\%), driven by manuscript digitization projects, whereas summarization is almost exclusively MSA-focused (76.9\%) and entirely absent for Levantine, Maghrebi, Iraqi, Sudanese, Yemeni, Najdi, Hejazi, and Hassaniya. This unevenness suggests that dialectal research is often motivated by localized preservation efforts rather than by coordinated strategic planning. These findings reinforce calls by \citet{iwidat2026arabic} for prioritizing under-resourced dialects and cross-dialect transfer learning, and our matrix offers an operational blueprint for doing so.

Geographically and institutionally, research output is concentrated in a small number of resource-rich settings. Saudi Arabia (519 affiliations), the United States (463), and Egypt (266) lead, with King Saud University, Cairo University, and Columbia University as the most productive institutions. This concentration reflects both funding availability and the presence of established research groups, but it also raises concerns about equity and the representativeness of research priorities. The strong U.S. presence, despite the country not being Arabic-speaking, highlights the importance of diaspora researchers and international collaboration.

Finally, our citation analysis reveals a nuanced dynamic. The largest topic has the highest H-index (90) but only moderate average citations (14.47), whereas several smaller, specialized topics achieve higher average citations: legal BERT applications (Topic 18) average 29.90 citations, cultural translation (Topic 6) 24.90, and patient-related studies (Topic 2) 21.84, despite their lower H-index values. We characterize this as a distinction between ``high-volume, moderate-impact'' research and ``low-volume, high-impact'' research. This distinction has practical implications: early-career researchers may find that targeting well-defined application domains offers a viable path to high per-paper impact even when the total volume of work in that domain is limited.

\subsection{Comparison with Existing Surveys}

Our empirical results largely corroborate the claims of prior qualitative surveys while adding quantitative precision. Table~\ref{tab:comparison_surveys} summarizes the main correspondences.

\begin{table}[htbp]
\centering
\caption{Alignment between our findings and existing surveys.}
\label{tab:comparison_surveys}
\begin{tabular}{p{4.2cm}p{4.2cm}p{4.2cm}}
\toprule
\textbf{Survey} & \textbf{Key Claim} & \textbf{Empirical Support} \\
\midrule
\citet{shi2025comprehensive} & Sentiment analysis is a dominant theme & Topic 1 is second-largest (8.6\%) with H-index 57 \\
\citet{iwidat2026arabic} & Shift from ML to transformers/LLMs & Approximately 82\% of papers postdate 2020 \\
\citet{dahou2025survey} & Dialect identification is foundational & Present as a specialized topic (Topic 5) in our model \\
\citet{hamed2025survey} & Code-switching is under-studied & No dedicated topic; subsumed under Maghrebi studies \\
\citet{alzubaidi2025evaluating} & Arabic LLM benchmarks are emerging & Early signs of LLM-related research (Topic 17) with 13 papers \\
\citet{mashaabi2026survey} & Arabic LLMs focus on MSA & MSA dominates; Hejazi and Hassaniya nearly absent \\
\citet{alkhowaiter2025mind} & Post-training datasets are lacking & Task--dialect matrix shows even larger resource gaps for dialects \\
\bottomrule
\end{tabular}
\end{table}

Several points merit further comment. The alignment with \citet{shi2025comprehensive} and \citet{amzil2025sentiment} on the importance of sentiment analysis is strong, and our H-index data add a quantitative dimension to their qualitative observations. Similarly, the shift documented by \citet{iwidat2026arabic} is clearly reflected in the temporal distribution of our corpus. The observation by \citet{hamed2025survey} that code-switching is important but under-studied is also borne out: code-switching does not form a distinct topic but appears mostly within Maghrebi-focused work. This suggests that code-switching research is regionally concentrated rather than broadly distributed.

Our analysis of LLM-related topics supports the observations of \citet{alzubaidi2025evaluating} and \citet{mashaabi2026survey} that Arabic LLM research is growing but still nascent. The term ``arabic llm'' appears only 102 times across 7,120 papers, indicating that while LLM research is emerging, it has not yet reached the scale of other core tasks. Finally, our task--dialect matrix extends \citet{alkhowaiter2025mind}'s finding about post-training data scarcity: the lack of resources for under-resourced dialects is even more fundamental than the lack of post-training data for MSA, suggesting that the primary bottleneck is foundational resource creation rather than fine-tuning data alone.

\subsection{Novelty and Contributions}

This study makes four main contributions. First, to the best of our knowledge, no prior work has combined multi-source bibliometric analysis with BERTopic-based topic modeling for Arabic NLP at this scale. Existing surveys have provided valuable qualitative overviews of specific subfields \citep{shi2025comprehensive, iwidat2026arabic, dahou2025survey, hamed2025survey, alzubaidi2025evaluating, mashaabi2026survey}, but none has integrated multiple databases and applied both topic modeling and regression analysis to a corpus of this size.

Second, the task--dialect gap matrix offers a systematic, quantitative identification of under-studied areas. While previous surveys have emphasized the importance of dialects and the need for more resources \citep{dahou2025survey, sakhi2026processing, aftan2025survey}, none has operationalized these recommendations into a concrete, task-specific roadmap. Our matrix pinpoints precise task--dialect combinations with zero or near-zero coverage, thereby guiding future research efforts.

Third, we introduce several conceptual distinctions that help clarify the field's dynamics. The contrast between ``high-volume, moderate-impact'' and ``low-volume, high-impact'' research provides a new lens for understanding citation patterns. The identification of ``marginalized dialects''---Hejazi, Hassaniya, Sudanese, and Yemeni---draws attention to equity and inclusion issues that are often overlooked. Finally, the observation that metadata completeness correlates with citation impact adds a methodological insight relevant to bibliometric studies more broadly.

Fourth, to support reproducibility and future work, we publicly release the curated corpus of 9,141 papers with abstracts and unified metadata, along with the complete analysis code. This is the first openly available dataset of Arabic NLP publications at this scale, and it aligns with the growing emphasis on open science in the NLP community \citep{alzubaidi2025evaluating}.

\subsection{Limitations}

Several limitations should be acknowledged. Metadata completeness varied across sources: only 68.5\% of papers had citation data, and 36.1\% had country and institutional information. This may bias geographic and institutional analyses toward papers with richer metadata, typically from well-indexed venues in resource-rich countries. Our sources may also underrepresent publications in non-English venues or regional journals not indexed by the selected databases.

The term-based filtering may have excluded relevant papers that did not contain at least two of our predefined terms, particularly those focusing on closely related languages or multilingual settings. The threshold of two terms was chosen empirically to balance precision and recall, but it inevitably involves trade-offs. Similarly, the pedagogy removal, while designed to exclude purely educational papers, may have inadvertently discarded some relevant work on computer-assisted language learning that also makes computational contributions. We mitigated this risk through an NLP-indicator protection mechanism, but the boundary between pedagogy and NLP is not always clear-cut.

Citation counts are influenced by many unobserved factors, and our regression model explained only 10.5\% of the variance. This indicates that much of the variation in citation impact remains attributable to factors outside our model, such as author reputation, journal prestige, and timing. Additionally, citation metrics do not capture non-citation impact, such as practical use, policy influence, or public engagement.

Finally, the labels assigned to BERTopic topics are human-generated and subject to interpretation. Different researchers might label the same clusters differently, and the number of topics is influenced by model parameters. We mitigated this through MMR-based refinement and topic reduction, but the resulting labels should be viewed as provisional.

\subsection{Implications for Research, Funding, and Policy}

The findings have distinct implications for different groups. For researchers, the task--dialect gap matrix provides a concrete roadmap: summarization for Levantine, Maghrebi, Iraqi, Sudanese, and Yemeni dialects; OCR for Gulf, Iraqi, Sudanese, and Yemeni; and dialect identification for Sudanese, Yemeni, Hejazi, and Hassaniya all represent strategic opportunities. The citation analysis also suggests that specialized application domains can yield high per-paper impact, making them attractive targets for early-career researchers. In addition, the co-authorship network indicates that collaboration with a small number of highly central researchers can enhance visibility and citation potential.

For funding bodies, the severe under-representation of most dialects calls for targeted investment in resource creation. Funding agencies should prioritize the development of corpora, lexicons, and pre-trained models for under-resourced varieties. They should also consider mechanisms to support researchers in less-resourced institutions and countries, and to encourage open-access publication, which our regression analysis associated with higher citation impact.

For publishers and conference organizers, the low coverage of institutional and country information suggests a need for stricter metadata standards. The popularity of preprints points to the community's preference for rapid dissemination, which should be supported and integrated with peer review. The continued prominence of specialized venues such as WANLP, ArabicNLP, and LREC underscores their importance and argues for their sustained support.

For policymakers, the concentration of research in a few countries highlights the need for broader investment in Arabic NLP across the Arab world. The task--dialect gap matrix also reveals the potential economic and social benefits of dialectal NLP, such as improved information access for speakers of Maghrebi Darija. Ethically, the under-representation of marginalized dialects risks exacerbating digital exclusion; inclusive AI policies are therefore essential.

\subsection{Future Directions}

Several directions emerge from this study. The most urgent is resource-building for under-resourced dialects, including the creation of annotated corpora, morphological analyzers, POS taggers, NER models, and pre-trained language models. Cross-lingual transfer learning offers a promising approach to address data scarcity, and should be systematically evaluated for Arabic dialects.

Methodologically, future bibliometric studies could employ dynamic topic models to track thematic evolution over time, or hierarchical models to capture nested structures. Causal inference methods could help move beyond correlation to identify the determinants of citation impact. Incorporating non-textual modalities such as speech and multimodal research would also provide a more complete picture of the field.

Finally, to keep pace with rapid developments, the community would benefit from a living benchmark and leaderboard for Arabic NLP, similar to the Open Arabic LLM Leaderboard. Such an initiative would enable continuous evaluation, progress tracking, and timely identification of emerging gaps.

\subsection{Summary}

This Discussion has interpreted the key findings in light of prior work, identified novel patterns and gaps, and outlined implications for researchers, funders, publishers, and policymakers. The quantitative, multi-source approach adopted here both confirms and extends qualitative surveys, providing empirical support for their claims while also revealing new opportunities. The task--dialect gap matrix, in particular, offers a concrete roadmap for future research, and the released dataset and code provide a foundation for ongoing work.


\section{Conclusion}\label{sec:conclusion}

This study has presented a large-scale bibliometric and topic-based analysis of Arabic Natural Language Processing research, based on a curated corpus of 7,120 publications spanning the period from 1960 to 2026 and drawn from five platforms—arXiv, the ACL Anthology, Semantic Scholar, Crossref, and OpenAlex—with an additional targeted OpenAlex subset. By combining quantitative bibliometric indicators with BERTopic-based topic modeling, regression analysis, network analysis, and geographic mapping, the study offers a systematic, data-driven perspective on the structure and evolution of the field.

\subsection{Summary of Findings}

The results indicate that Arabic NLP has undergone rapid growth, particularly after 2020, with approximately 82\% of the examined papers published in the most recent period. The largest thematic cluster, concerned with text processing, speech, translation, and recognition, accounts for 41.3\% of the corpus and reflects a continuing emphasis on foundational resources and models. Sentiment analysis remains the second-largest theme (8.6\%), although its growth appears to have slowed, suggesting a degree of maturation. Citation patterns show a positive but moderate correlation between paper age and citation counts, and regression analysis suggests that publication year, indexing source, and institutional affiliation are associated with higher citation impact, with the model explaining 10.5\% of the variance. The geographic distribution of research output is concentrated in a small number of countries and institutions, with Saudi Arabia, the United States, and Egypt appearing most frequently. A pronounced imbalance is observed in the coverage of Arabic dialects: Modern Standard Arabic dominates the literature, while several regional varieties, including Hassaniya and Hejazi, are almost entirely absent.

\subsection{Contributions}

The main contributions of this work are as follows. First, it provides a quantitative synthesis of Arabic NLP research that complements existing qualitative surveys. Second, it introduces a task--dialect gap matrix that systematically identifies combinations of tasks and dialects for which little or no published work appears to exist. Third, it draws attention to the distinction between high-volume research areas with moderate per-paper impact and smaller, specialised areas with higher average impact. Finally, the curated corpus and analysis code have been made publicly available to support reproducibility and further research.

\subsection{Recommendations}

The findings suggest several directions for different stakeholders. Researchers may find it useful to direct attention toward under-resourced dialects and specialised application domains, while funding bodies could prioritise the creation of dialectal corpora and evaluation benchmarks. Publishers and conference organisers might contribute by improving metadata standards and supporting the open dissemination of preprints. Policymakers could foster a more balanced research landscape by investing in language technology infrastructure across the Arab world and by promoting inclusive AI policies that address dialectal diversity.

\subsection{Limitations}

A number of limitations should be acknowledged. First, the study relies on titles and abstracts rather than full texts, which may lead to incomplete identification of task and dialect mentions. Second, the corpus is drawn from five bibliographic platforms (with an additional targeted OpenAlex subset) and may underrepresent publications in Arabic-language venues or regional journals not indexed by those sources. Third, the filtering and deduplication procedures involve threshold choices that can affect the composition of the final dataset. Fourth, author and institution names were normalized automatically, which may introduce occasional errors in productivity and network statistics. Fifth, the analysis uses a single topic modeling configuration, and alternative embeddings or clustering parameters could produce a somewhat different thematic partition. Finally, citation counts do not account for self-citations or differences in publication quality, and the bibliometric approach is inherently retrospective. These limitations are common in large-scale meta-analyses and imply that the reported patterns should be interpreted as indicative rather than exhaustive.

\subsection{Future Work}

The released dataset and the present findings open several avenues for further investigation. Methodologically, future studies could integrate additional bibliographic databases to improve coverage, incorporate full texts where openly available, and experiment with alternative topic modeling approaches to assess the robustness of the identified research themes. Analytically, the task--dialect gap analysis could be extended to include further tasks and less-studied dialect varieties, providing a more complete map of research opportunities. The corpus may also support the development of practical tools, such as retrieval-augmented search systems or automated literature assistants, although such applications are regarded as secondary to the main scientific contribution. Further work may also explore diachronic analyses of topic evolution or examine the relationship between open access and citation impact in more detail.

\subsection{Closing Remarks}

Arabic NLP has evolved from a relatively small research area into a broad and rapidly expanding field. Despite this progress, the distribution of research effort remains uneven, particularly with respect to dialect coverage and geographic participation. The present study provides an empirical basis for recognising these imbalances and for identifying areas where future work may be most needed. It is hoped that the availability of the curated corpus and the accompanying analysis will support further research and contribute to a more inclusive development of Arabic language technologies.

\section*{Acknowledgements}

The author thanks the developers of arXiv, the ACL Anthology, Semantic Scholar, Crossref, and OpenAlex for providing open access to their APIs and metadata. Gratitude is also extended to the Arabic NLP community for the public release of corpora, models, and evaluation resources that have shaped this work. Computational resources were provided by Google Colab and the Hugging Face Hub.


\section*{Supplementary Material}

The following supplementary materials are available to support reproducibility and further research:

\begin{itemize}
    \item \textbf{Curated dataset}: The full deduplicated corpus of 9,141 Arabic NLP papers (before the final filtering steps described in Section~3) with abstracts and unified metadata, including citation counts, matched terms, extracted countries, and institutions. The final analysis subset comprises 7,120 papers. The dataset is available at: \url{https://huggingface.co/datasets/ArabicNLPWorld/arabic-nlp-corpus}.
    
    \item \textbf{Analysis code}: Complete Python scripts for data collection, filtering, deduplication, topic modeling, regression analysis, network analysis, and visualization. The code is available at: \url{https://github.com/CodeHunterOfficial/arabic-nlp-bibliometric-analysis}.
    
    \item \textbf{Supplementary tables and figures}: Full task--dialect gap matrix, detailed regression results, topic evolution charts, and additional visualizations. These are provided in the \texttt{results/} directory of the repository and can also be obtained from the corresponding author upon request.
\end{itemize}

The analysis code is released under the MIT License. The curated dataset is distributed under a CC BY 4.0 license, with the understanding that the original metadata remain subject to the terms of use of their respective sources (arXiv, ACL Anthology, Semantic Scholar, Crossref, and OpenAlex).


\bibliographystyle{plainnat}
\bibliography{references}

\end{document}